\documentclass[lettersize,journal]{IEEEtran}
\usepackage{amsmath,amsfonts}
\usepackage{algorithmic}
\usepackage{algorithm}
\usepackage{array}
\usepackage[caption=false,font=normalsize,labelfont=sf,textfont=sf]{subfig}
\usepackage{textcomp}
\usepackage{stfloats}
\usepackage{url}
\usepackage{verbatim}
\usepackage{graphicx}
\usepackage{cite}
\usepackage{makecell}
\usepackage{multirow}
\usepackage[table]{xcolor}
\allowdisplaybreaks[4]
\usepackage{amssymb}

\begin{document}

\title{Online Learning Under A Separable Stochastic Approximation Framework}

\author{Min Gan,~\IEEEmembership{Senior Member,~IEEE,}~Xiang-xiang Su, Guang-yong Chen
\thanks{M. Gan, X.-x. Su, G.-Y. Chen are with the College of Computer and Data Science, Fuzhou University, Fuzhou 350116, China(e-mail: aganmin@aliyun.com; cgykeda@mail.ustc.edu.cn; ).}
\thanks{Manuscript received May **, 2023; revised ****.}
\thanks{$\dagger$ https://github.com/keeplearningkeep/SepSA.git}
}

\markboth{Journal of \LaTeX\ Class Files,~Vol.~14, No.~8, August~2021}%
{Shell \MakeLowercase{\textit{et al.}}: A Sample Article Using IEEEtran.cls for IEEE Journals}


\maketitle

\begin{abstract}
  We propose an online learning algorithm for a class of machine learning models under a separable stochastic approximation framework. The essence of our idea lies in the observation that certain parameters in the models are easier to optimize than others. In this paper, we focus on models where some parameters have a linear nature, which is common in machine learning. In one routine of the proposed algorithm, the linear parameters are updated by the recursive least squares (RLS) algorithm, which is equivalent to a stochastic Newton method; then, based on the updated linear parameters, the nonlinear parameters are updated by the stochastic gradient method (SGD). The proposed algorithm can be understood as a stochastic approximation version of block coordinate gradient descent approach in which one part of the parameters is updated by a second-order SGD method while the other part is updated by a first-order SGD. Global convergence of the proposed online algorithm for non-convex cases is established in terms of the expected violation of a first-order optimality condition. Numerical experiments have shown that the proposed method accelerates convergence significantly and produces more robust training and test performance when compared to other popular learning algorithms. Moreover, our algorithm is less sensitive to the learning rate and outperforms the recently proposed \texttt{slimTrain} algorithm. The code has been uploaded to GitHub for validation$\dagger$.
  
 
\end{abstract}


\section{Introduction}
\IEEEPARstart{M}{achine} learning tasks are often reduced to minimizing an \textit{expected risk function} which may be defined as follows \cite{ref1, ref2}:
\begin{equation}
  f(\textbf{w})=\mathbb{E}_{\xi} F(\textbf{w},\xi)=\int F(\textbf{w},\xi)dP(\xi)
\end{equation}
where $\mathbb{E}$ is the expectation operator, the vector $\textbf{w}$ represents the optimization variable in the learning system, $\xi$ is a random vector having probability distribution $P$, and $F(\textbf{w},\xi)$ is the user-defined real-valued (continuously differentiable) loss function that measures the performance of the learning system with parameter $\textbf{w}$ under the circumstances described by $\xi$.

The optimal parameter $\textbf{w}^{*}$ is then determined by 
\begin{equation}
  \textbf{w}^{*}=\arg \min_{\textbf{w}} f(\textbf{w}).
\end{equation}

Unfortunately, the problem (2) is usually intractable since the probability distribution $P$ is unknown and therefore the expectation in (1) can not be evaluated. One straightforward way around this is to approximate the expectation with sample means by using a finite training set of independent observations $\xi_1$, $\xi_2$, $\cdots$, $\xi_m$:
\begin{equation}
  f(\textbf{w})\approx f_m(\textbf{w}) =\frac{1}{m} \sum_{i=1}^{m}F(\textbf{w},\xi_i).
\end{equation}

The \textit{batch} or \textit{offline} learning methods minimize the \textit{empirical risk function} (3) 
\begin{equation}
  \min_{\textbf{w}} f_m(\textbf{w})=\frac{1}{m} \sum_{i=1}^{m }F(\textbf{w},\xi_i)
\end{equation}
using the entire training data at once. In this way, numerous deterministic (numerical) optimization methods \cite{ref3} including first-order and second-order techniques can be applied to minimize (3). Usually, function (3) is highly nonlinear and non-convex (e.g., the neural networks) \cite{ref4} and has many local minima \cite{ref5,ref6} which makes it hard to solve.

Fortunately, many optimization problems in machine learning are "structured", i.e., the optimization with respect to some of the parameters in $\textbf{w}$ is convex, making them easier to be solved than others. For instance, when training feedforward neural networks with linear neurons in the output layer and using the mean squared error as the loss function, the optimization problem of the weights in the output layer is convex and can be easily solved by the linear least squares method. Similarly, when the output layer of a neural network uses sigmoid neurons and cross-entropy is the loss function, the optimization problem of the weights in the last layer is also convex, and it can be efficiently solved by a Newton-Raphson iterative scheme \cite{ref7}. The same holds true for certain computer vision problems such as low-rank matrix factorization \cite{ref8} and non-negative matrix factorization \cite{ref9}. This class of optimization problems are usually called \textit{separable}. 

Taking advantage of the special structure of the problem, one can develop very efficient algorithms. To explain this problem mathematically, we suppose the optimization variable $\textbf{w}$ can be partitioned into 
\begin{equation}
   \textbf{w}= \begin{pmatrix}
      \boldsymbol{\alpha} \\
      \boldsymbol{\theta}
      \end{pmatrix} 
\end{equation}
in such a way that the subproblem
\begin{equation}
  \min_{\boldsymbol{\alpha}} f_m(\boldsymbol{\alpha},\boldsymbol{\theta})
\end{equation}
is easy to solve analytically or numerically for any fixed $\boldsymbol{\theta}$.

Apparently, subproblem (6) is easier to be solved if it is convex. The simplest case is the linear least squares problems with closed form of solutions. Denote the solution of (6) as $\boldsymbol{\alpha}(\boldsymbol{\theta})$ and replace $\boldsymbol{\alpha}$ in the original functional in (4), the minimization problem (4) then becomes
\begin{equation}
  \min_{\boldsymbol{\theta}} \psi(\boldsymbol{\theta})
\end{equation}
where 
\begin{equation}
  \psi(\boldsymbol{\theta})=f_m(\boldsymbol{\alpha}(\boldsymbol{\theta}),\boldsymbol{\theta}).
\end{equation}

Let $\textbf{w}\in\Re^{n}$, $\boldsymbol{\alpha}\in\Re^{p}$, $\boldsymbol{\theta}\in\Re^{q}$, and $n=p+q$. It can be seen that we now turn an $n$ dimensional minimization problem (4) into a $q$ dimensional one (7) at the cost of the computation of $\boldsymbol{\alpha}(\boldsymbol{\theta})$ for every evaluation of the objective function. When the subproblem (6) is a linear least squares problem and the $f_m(\textbf{w})$ is nonlinear least squares, Golub \& Pereyra \cite{ref10} referred to the problem as the \textit{separable nonlinear least squares} (SNNLS) problems which have been extensively studied \cite{ref11, ref12, ref13, ref14, ref15, ref16, ref17}. The advantages of this separated paradigm, as opposed to that of using a general nonlinear optimization to estimate all parameters directly, are that i) the reduced problems are better conditioned \cite{ref18}; ii) in general, fewer iterations are required to convergence; iii) it works in a reduced parameter space, therefore requiring fewer initial guesses.

Theoretically, in many situations, we can conveniently calculate the gradient or even the (approximated) Hessian matrix of the resulting optimization problem (7), which is then can be solved by deterministic methods. However, in the age of big data, the dimension of $\textbf{w}$ in many machine learning tasks is huge, and the size of dataset is massive. Batch learning algorithms require significant computational resources in such situations, and therefore numerical evaluation of the gradient or Hessian is intractable. 

An alternative to the approach (3) of replacing expectations with sample means is to directly optimize the expected risk by applying \textit{stochastic approximation} schemes. Because this paradigm approximates the gradient of the expected risk with a single sample (or a small, randomly chosen batch of samples), it alleviates the computational complexity problem when using batch optimization methods. Stochastic approximation algorithms are also referred to as \textit{recursive identification} in the field of system identifiaction \cite{ref19}, \textit{sequential estimation} in statistics, \textit{adaptive algorithms} in signal processing, and \textit{online learning} in machine learning \cite{ref20}. Currently, the most popular stochastic approximation algorithms in machine learning are stochastic graidient decent (SGD) and their variants, which have become the main workhorse for training NN models. Bottou \& LeCun \cite{ref20} argued that suitably designed online learning algorithms asymptotically outperform any batch learning algorithm when datasets grow to practically infinite sizes.  This naturally motivates us to develop algorithms relying on stochastic gradients to solve the machine learning problems (1) whose parameters are separable.   

While much effort \cite{ref14, ref15, ref16, ref17, ref18, ref21, ref22, ref23, ref24, ref25, ref26, ref27, ref28} has been devoted to the batch learning algorithms for separable nonlinear optimization in the last several decades, few works of online learning algorithms for this type of optimization can be found in the literature. We list the related work of online learning for this topic to best of our knowledge. Asirvadam \textit{et al.} \cite{ref29} proposed hybrid training algorithms for NNs where they combined nonlinear recursive optimization of hidden layer nonlinear weights with recursive least squares (RLS) optimization of linear output layer weights. These strategies optimized the linear and nonlinear parameters alternatively. Gan \textit{et al.} \cite{ref30} proposed a recursive variable projection algorithm that considered the coupling between the variables. Recently, Chen \textit{et al.} \cite{ref31} improved the algorithm of \cite{ref30} by introducing an embedded point iteration step. These methods applied second-order recursive algorithms to both the linear and nonlinear parameters. However, obtaining curvature information is computationally heavy, especially when there are many nonlinear parameters in the models, such as in the case of deep neural networks. In view of this point, Newman \textit{et al.} \cite{ref32} and Chen \textit{et al.} \cite{ref33} used first-order SGD to update the nonlinear parameters based on the principle of variable projection \cite{ref10}.

In this paper, we present a general perspective on the stochastic separable optimization problem, i.e., in equation (1) $\textbf{w}$ can be partitioned into two parts $\textbf{w}=(\boldsymbol{\alpha}^T,\boldsymbol{\theta}^T)^T$ and $f(\boldsymbol{\alpha}, \boldsymbol{\theta})$ is convex when $\boldsymbol{\theta}$ is fixed.  In such cases, we can design more efficient updating strategies for the convex problem, forming the idea of the separable stochastic approximation framework. Our focus in this paper is on situations where some of the parameters in the model appear linear, and the loss function is taken as the sum of squared errors. An online algorithm is then derived from the separable stochastic approximation framework. In one routine of the proposed algorithm, the linear parameters are updated using the recursive least squares algorithm, which is equivalent to a stochastic Newton method; then, based on the updated linear parameters, the nonlinear parameters are updated by the stochastic gradient method (SGD). This proposed algorithm can be understood as a stochastic approximation version of block coordinate descent approach in which part of the parameters are optimized by a second-order SGD method and the other part is optimized by the first-order SGD. Numerical experiments demonstrate the efficiency and effectiveness of the proposed approach.

The contribution of this paper are as follows.
\begin{itemize}
  \item We introduce a class of stochastic separable optimization problems and proposed a separable stochastic approximation framework for solving them.\\
  \item Under the separable stochastic approximation framework, we propose an online algorithm, named SepSA, for solving the stochastic separable nonlinear least squares problem. Global convergence of the proposed online algorithm for non-convex cases is established.\\
  \item We performed extensive experiments on two classic tasks, regression and classification, to compare the performance of the SepSA with that of other widely-used algorithms. The experimental results demonstrate that the SepSA algorithm exhibits notable advantages, such as faster convergence speed, less sensitivity to learning rate, and greater suitability for online learning. \end{itemize}

Our paper is organized as follows: In Section II, we present the separable stochastic approximation framework and propose an online algorithm for solving the stochastic separable nonlinear least squares problem. In Section III, we analyze the convergence of the proposed online algorithm for the stochastic separable nonlinear least squares problem in terms of the expected violation of a first-order optimality condition. In Section IV, we present experimental results that compare our proposed algorithm with other commonly used algorithms.

\section{Separable Stochastic Approximation Framework and Online Learning Algorithm}
As discussed in the Introduction, the optimization problem we studied in this paper is separable, i.e., the variable $\textbf{w}$ can be partitioned into $\textbf{w}=(\boldsymbol{\alpha}^T,\boldsymbol{\theta}^T)^T$ and $f(\boldsymbol{\alpha},\boldsymbol{\theta})$ is convex when we fix $\boldsymbol{\theta}$. Such situations are very common in machine learning. Then, the proposed separable stochastic approximation framework is given below.
\begin{algorithm}[h]
\caption{Separable Stochastic Approximation Framework}
\begin{algorithmic}
\STATE Initialize $\boldsymbol{\alpha}_0,\boldsymbol{\theta}_0$
\STATE \textbf{for} $k=1,2, \cdots$
\STATE \hspace{0.5cm} Update $\boldsymbol{\alpha}_{k}$ based on $F(\boldsymbol{\alpha},\boldsymbol{\theta}_{k-1},\xi_k)$ using efficient stochastic convex optimization algorithm. 
\STATE \hspace{0.5cm} Update $\boldsymbol{\theta}_{k}$ based on $F(\boldsymbol{\alpha}_{k},\boldsymbol{\theta},\xi_k)$ using stochastic gradient decent algorithm.
\STATE \textbf{end}
\end{algorithmic}
\end{algorithm}

For stochastic convex optimization problems \cite{ref34, ref35}, it is usually relatively easier to design efficient stochastic approximation algorithms \cite{ref36, ref37, ref38, ref39, ref40}. Thus, the separation in the \textit{Separable Stochastic Approximation Framework} allows us to leverage the advances made for stochastic convex optimization. In this paper, we focus on the situations where the models have some linear parameters and the loss function is the expected squared error. We plan to explore other situations in the future work. 

Under the machine learning settings, without of loss generality, we assume that $\xi$ is the random instance consisting of an input-output pair $(\boldsymbol{x},y)$ where $\boldsymbol{x} \in \Re^d$ represents the input of the learning system and $y \in \Re$ is the target output  and that the underlying models of the machine learning tasks have the following form:
\begin{equation}
  \eta(\boldsymbol{x};\boldsymbol{\alpha},\boldsymbol{\theta})=\boldsymbol{\alpha}^T \boldsymbol{h}(\boldsymbol{x};\boldsymbol{\theta})
\end{equation}
where $\boldsymbol{h}$ can be regraded as a (nonlinear) feature extractor. The loss function is the squared error
\begin{equation}
  F(\boldsymbol{\alpha},\boldsymbol{\theta},\xi)=\frac{1}{2}\left(y-\eta(\boldsymbol{x};\boldsymbol{\alpha},\boldsymbol{\theta})\right)^2.
\end{equation}
Now the minimization problem becomes
\begin{equation}
  \min_{\boldsymbol{\alpha},\boldsymbol{\theta}} \mathbb{E}_{\xi} F(\boldsymbol{\alpha},\boldsymbol{\theta},\xi)
\end{equation}
which is referred to as \textit{Stochastic Separable Nonlinear Least Squares} (SSNLS).

It is apparently that (11) is a stochastic linear least squares problem which is convex if $\boldsymbol{\theta}$ is fixed. In this paper, we follow the way of Ljung \& S$\ddot{o}$derstr$\ddot{o}$m \cite{ref19} to solve this stochastic convex optimization problem. The scheme is actually a stochastic Newton algorithm which employs the following iterations of the form
\begin{equation}
  \boldsymbol{\alpha}_k = \boldsymbol{\alpha}_{k-1} - \gamma_k \boldsymbol{H}^{-1}_{k}\boldsymbol{g}_k
\end{equation}
where $\gamma_k$ is a step size that is typically required to asymptotically reduce to zero for convergence, $\boldsymbol{H}_k$ is a symmetric positive definite approximation to the Hessian matrix $\nabla^2_{\boldsymbol{\alpha}} f(\boldsymbol{\alpha}_{k-1},\boldsymbol{\theta}_{k-1})$ at the $k$th iteration, $\boldsymbol{g}_k$ is a stochastic estimate of the gradient with respect to $\boldsymbol{\alpha}$
\begin{equation}
\begin{aligned}
  \boldsymbol{g}_k &= \nabla_{\boldsymbol{\alpha}} F(\boldsymbol{\alpha}_{k-1},\boldsymbol{\theta}_{k-1},\xi_k) \\
  &=\boldsymbol{h}(\boldsymbol{x}_k;\boldsymbol{\theta}_{k-1})\left(\eta(\boldsymbol{x}_k;\boldsymbol{\alpha}_{k-1},\boldsymbol{\theta}_{k-1})-y_k\right)\\
  &=\boldsymbol{h}(\boldsymbol{x}_k;\boldsymbol{\theta}_{k-1})\left(\boldsymbol{\alpha}^T_{k-1} \boldsymbol{h}(\boldsymbol{x}_k;\boldsymbol{\theta}_{k-1})-y_k\right),
 \end{aligned}
\end{equation}
and $\xi_k$ is a realization of $\xi$ at the $k$th iteration.

Suppose that the approximation $\boldsymbol{H}_k$ of the Hessian $\nabla^2_{\boldsymbol{\alpha}} f(\boldsymbol{\alpha}_{k-1},\boldsymbol{\theta}_{k-1})$ can be constructed from previous samples and note that 
\begin{equation}
  \nabla^2_{\boldsymbol{\alpha}} f(\boldsymbol{\alpha},\boldsymbol{\theta})=\mathbb{E}_\xi \left[ \boldsymbol{h}(\boldsymbol{x};\boldsymbol{\theta})\boldsymbol{h}^T(\boldsymbol{x};\boldsymbol{\theta}) \right] 
\end{equation}
is independent of $\boldsymbol{\alpha}$, thus the Hessian can be determined as the solution $\boldsymbol{H}$ of the equation
\begin{equation}
  \mathbb{E}_\xi \left[ \boldsymbol{h}(\boldsymbol{x};\boldsymbol{\theta})\boldsymbol{h}^T(\boldsymbol{x};\boldsymbol{\theta}) - \boldsymbol{H} \right]=0.
\end{equation}

Applying the Robbins-Monro procedure \cite{ref41} to (15), we obtain
\begin{equation}
  \boldsymbol{H}_k=\boldsymbol{H}_{k-1} + \gamma_k \left[ \boldsymbol{h}(\boldsymbol{x}_k;\boldsymbol{\theta}_{k-1})\boldsymbol{h}^T(\boldsymbol{x}_k;\boldsymbol{\theta}_{k-1}) - \boldsymbol{H}_{k-1} \right].
\end{equation}

The iterations (12) and (16) give a complete stochastic Newton update for the variable $\boldsymbol{\alpha}$ with fixed $\boldsymbol{\theta}$. Let $\gamma_k=\frac{1}{k}$, we can see that it coincides with the well-known \textit{recursive least squares} (RLS) formula. If the dimension of $\boldsymbol{H}_k$ is large, the computation of inverse can be onerous. To avoid the inverse, the matrix inversion lemma is applied to (12) and (16), and we consequently derived the RLS algorithm which is given by (denote $\boldsymbol{B}_{k}=\frac{1}{k}\boldsymbol{H}^{-1}_{k}$) 
\begin{align}
 \boldsymbol{\alpha}_k & = \boldsymbol{\alpha}_{k-1} - \boldsymbol{B}_{k}\boldsymbol{g}_k, \\
\boldsymbol{B}_k & = \boldsymbol{B}_{k-1} - \frac{\boldsymbol{B}_{k-1}\boldsymbol{h}(\boldsymbol{x}_k;\boldsymbol{\theta}_{k-1})\boldsymbol{h}^T(\boldsymbol{x}_k;\boldsymbol{\theta}_{k-1})\boldsymbol{B}_{k-1}}{1+\boldsymbol{h}^T(\boldsymbol{x}_k;\boldsymbol{\theta}_{k-1})\boldsymbol{B}_{k-1}\boldsymbol{h}(\boldsymbol{x}_k;\boldsymbol{\theta}_{k-1})}.
\end{align}

For nonlinear variable $\boldsymbol{\theta}$ in (9), we can adopt the classical SGD or its variants such as AdaGrad \cite{ref42} and Adam \cite{ref43}. The update may be formulated as
\begin{equation}
  \boldsymbol{\theta}_k = \boldsymbol{\theta}_{k-1} + \beta_{k}\boldsymbol{p}_k,
\end{equation}
where $\beta_{k}$ is the appropriate step size and $\boldsymbol{p}_k$ is the search direction with respect to $\boldsymbol{\theta}$ that may be computed based on $F(\boldsymbol{\alpha}_k, \boldsymbol{\theta}_{k-1}, \xi_k)$ and other previous variables. For the classical SGD, $\boldsymbol{p}_k=-\nabla_{\boldsymbol{\theta}} F(\boldsymbol{\alpha}_k, \boldsymbol{\theta}_{k-1}, \xi_k)$.
Next, we summarize our proposed online learning algorithm (SepSA) for the stochastic separable nonlinear least squares as follows.
\begin{algorithm}[h]
\caption{Online algorithm for Stochastic Separable Nonlinear Least Squares}
\begin{algorithmic}
\STATE Initialize $\boldsymbol{\alpha}_0,\boldsymbol{\theta}_0, \boldsymbol{B}_0$
\STATE \textbf{for} $k=1,2, \cdots$
\STATE \hspace{0.5cm} Update $\boldsymbol{\alpha}_{k}$ using (17) and (18). 
\STATE \hspace{0.5cm} Update $\boldsymbol{\theta}_{k}$ using (19).
\STATE \textbf{end}
\end{algorithmic}
\end{algorithm}

Note that for simplicity we do not include regularization terms in the objective function and assume single-output in the model, however it is easy to extend the algorithm to the cases that involves regularization terms and multi-output. The algorithm described above also applies to the case of \textit{mini-batch} gradient decent. We have the following remarks about the Algorithm 2.

\textbf{Remark 1.} The proposed separable stochastic approximation framework and the specific Algorithm 2 for SSNLS problems can be regarded as a special case of the block stochastic gradient iteration algorithm \cite{ref44} which generalizes the SGD by updating all the block of variables in the Gauss-SeSeidel manner. However, we leverage the advances of the special structure of separable optimization problems, which apparently improves the performance of the algorithm. In \cite{ref44}, the authors demonstrated that block stochastic gradient algorithms offer another advantage: they can handle larger step sizes compared to regular stochastic gradient descent (SGD).\\

\textbf{Remark 2.} A relevant work to ours is \cite{ref32}, which proposes a stochastic approximation version of the variable projection approach (they call it \textbf{slimTrain}) for SNNLS problems. For the convenience of our discussion, we rewrite their optimization problem in the following
\begin{equation}
\begin{aligned}
  \min_{\boldsymbol{W},\boldsymbol{\theta}} \Phi(\boldsymbol{W},\boldsymbol{\theta}) = \mathbb{E}_\xi \frac{1}{2}\left\| \boldsymbol{W}\boldsymbol{h}(\boldsymbol{x};\boldsymbol{\theta})- \boldsymbol{y} \right \|_2^2 \\
  + \frac{\alpha}{2}\left\| \boldsymbol{L}\boldsymbol{\theta} \right \|_2^2 + \frac{\lambda}{2}\left\| \boldsymbol{W} \right \|_F^2, 
  \end{aligned}
\end{equation}
where $\boldsymbol{W}$ is the linear parameter matrix, $\xi$ is the random instance consisting of multi-input multi-output pair $(\boldsymbol{x},\boldsymbol{y})$, $\left\| \cdot \right \|_2$ and $\left\| \cdot \right \|_F$ are the $l_2$-norm and Frobenius norm, respectively, $\boldsymbol{L}$ is a user-defined operator, $\alpha$ and $\lambda$ are the regualarization parameters.

The \texttt{slimTrain} is to solve the reduced stochastic optimization problem
\begin{equation}
  \min_{\boldsymbol{\theta}} \Phi^{red}(\boldsymbol{\theta})=\Phi(\hat{\boldsymbol{W}}(\boldsymbol{\theta}),\boldsymbol{\theta}),
\end{equation}
where 
\begin{equation}
  \hat{\boldsymbol{W}}(\boldsymbol{\theta})=\arg \min_{\boldsymbol{W}} \mathbb{E} \frac{1}{2}\left\| \boldsymbol{W}\boldsymbol{h}(\boldsymbol{x};\boldsymbol{\theta})- \boldsymbol{y} \right \|_2^2 + \frac{\lambda}{2}\left\| \boldsymbol{W} \right \|_F^2.
\end{equation}
That is, for every iteration of updating $\boldsymbol{\theta}$ in (21), one has to solve a stochastic Tikhonvo-regularized linear least squares problem \textit{over the entire data space}, which is apparently impractical for large-scale problems. As stated in \cite{ref32}, a practical way to approximate $\hat{\boldsymbol{W}}(\boldsymbol{\theta})$ is the sample average approximation (SAA) approach, which needs a large number of samples. Although \texttt{slimTrain} used the so-called iterative sampled limited memory method to approximate $\hat{\boldsymbol{W}}(\boldsymbol{\theta})$, the computation burden is still heavy to obtain a satisfied approximation.

Conversely, in this paper we abandon the idea of eliminating linear parameters from the problem and have instead adopted the block coordinate gradient iteration approach. In this way, our proposed algorithm is easier to implement and computationally lighter at each iteration. It is not only applicable to stochastic programming in the form of (1) but also to deterministic problems in the form of (4) with a huge amount of training data.

\section{Convergence Analysis}
\subsection{Global Convergence for Nonconvex Case}
In this subsection, we analyze the convergence of Algorithm 2 under the setting that $f$ is nonconvex. Our analysis is similar to analyses in \cite{ref39, ref44}. The challenge of the analysis lies in the biased partial gradient and the approximated Hessian matrix. Without loss of generality, we assume that the formulas of updating $\boldsymbol\alpha$ and $\boldsymbol\theta$ are
\begin{align}
  \boldsymbol{\alpha}_k &= \boldsymbol{\alpha}_{k-1} - \gamma_k \boldsymbol{H}^{-1}_{k}\nabla_{\boldsymbol{\alpha}} F(\boldsymbol{\alpha}_{k-1},\boldsymbol{\theta}_{k-1},\xi_k),\\
  \boldsymbol{\theta}_k &= \boldsymbol{\theta}_{k-1} - \beta_{k}\nabla_{\boldsymbol{\theta}} F(\boldsymbol{\alpha}_k, \boldsymbol{\theta}_{k-1}, \xi_k).
\end{align}
Denote $\textbf{w}_k=(\boldsymbol\alpha_k;\boldsymbol\theta_k)$. We make the following assumptions for the analysis.

ASSUMPTION 1. $f(\textbf{w})$ \textit{is lower bounded, i.e.,} $f(\textbf{w})>-\infty$. \textit{The partial gradient of} $f$ \textit{is Lipschitz continuous and there is a uniform Lipschitz constant $L>0$ such as}
\begin{align}
  \left\| \nabla_{\boldsymbol{\alpha}} f(\textbf{w}) - \nabla_{\boldsymbol{\alpha}} f(\tilde{\textbf{w}}) \right\|_2 &\leq L\left\| \textbf{w}- \tilde{\textbf{w}} \right\|_2, \forall \textbf{w}, \tilde{\textbf{w}};\\
  \left\| \nabla_{\boldsymbol{\theta}} f(\textbf{w}) - \nabla_{\boldsymbol{\theta}} f(\tilde{\textbf{w}}) \right\|_2 &\leq L\left\| \textbf{w}- \tilde{\textbf{w}} \right\|_2, \forall \textbf{w}, \tilde{\textbf{w}}.
\end{align}

ASSUMPTION 2. \textit{For every iteration} $k$, 
\begin{equation}
  \mathbb{E} \left\| \textbf{w}_k \right\|_2^2 \leq \rho^2,
\end{equation}
\textit{where $\rho$ is a constant}.

\textbf{Remark 3}. By the Lipschitz continuity of $\nabla_{\boldsymbol{\alpha}} f(\textbf{w})$, we have 
\begin{equation}\notag
\begin{aligned}
  &\left\| \nabla_{\boldsymbol{\alpha}} f(\textbf{w}_{k}) \right\|_2^2 \\
  & \leq 2\left\| \nabla_{\boldsymbol{\alpha}} f(\textbf{w}_{k})-\nabla_{\boldsymbol{\alpha}} f(\boldsymbol{0})\right\|_2^2 + 2\left\| \nabla_{\boldsymbol{\alpha}} f(\boldsymbol{0})\right\|_2^2 \\
  & \leq 2L^2 \left\| \boldsymbol{\textbf{w}}_{k} \right\|_2^2 + 2\left\| \nabla_{\boldsymbol{\alpha}} f(\boldsymbol{0})\right\|_2^2.
\end{aligned}
\end{equation}
Thus, by Assumption 2, we obtain
\begin{equation}
  \begin{aligned}
    & \mathbb{E} \left\| \nabla_{\boldsymbol{\alpha}} f(\textbf{w}_{k}) \right\|_2^2  \leq 2L^2 \rho^2 + 2\left\| \nabla_{\boldsymbol{\theta}} f(\boldsymbol{0})\right\|_2^2.
  \end{aligned}
\end{equation}
Similarly, 
\begin{align}
    \mathbb{E} \left\| \nabla_{\boldsymbol{\theta}} f(\textbf{w}_{k}) \right\|_2^2  \leq 2L^2 \rho^2 + 2\left\| \nabla_{\boldsymbol{\theta}} f(\boldsymbol{0})\right\|_2^2.
  \end{align}
By the Lipschitz continuity of $\nabla_{\boldsymbol{\theta}} f(\textbf{w})$, we have 
\begin{equation}
\begin{aligned}
  &\left\| \nabla_{\boldsymbol{\theta}} f(\boldsymbol{\alpha}_{k},\boldsymbol{\theta}_{k-1}) \right\|_2^2 \\
  & \leq 2\left\| \nabla_{\boldsymbol{\theta}} f(\boldsymbol{\alpha}_{k},\boldsymbol{\theta}_{k-1})-\nabla_{\boldsymbol{\theta}} f(\boldsymbol{0})\right\|_2^2 + 2\left\| \nabla_{\boldsymbol{\theta}} f(\boldsymbol{0})\right\|_2^2 \\
  & \leq 2L^2 \left\| \boldsymbol{\alpha}_{k};\boldsymbol{\theta}_{k-1} \right\|_2^2 + 2\left\| \nabla_{\boldsymbol{\theta}} f(\boldsymbol{0})\right\|_2^2.
\end{aligned}
\end{equation}
Thus, by Assumption 2, we obtain
\begin{equation}
  \begin{aligned}
    & \mathbb{E} \left\| \nabla_{\boldsymbol{\theta}} f(\boldsymbol{\alpha}_{k},\boldsymbol{\theta}_{k-1}) \right\|_2^2  \leq 4L^2 \rho^2 + 2\left\| \nabla_{\boldsymbol{\theta}} f(\boldsymbol{0})\right\|_2^2.
  \end{aligned}
\end{equation}
That is, Assumption 2 together with the partial gradient Lipschitz continuity of $f$ in Assumption 1 implies the boundedness of $\mathbb{E} \left\| \nabla_{\boldsymbol{\alpha}} f(\textbf{w}_{k}) \right\|_2^2$, $\mathbb{E} \left\| \nabla_{\boldsymbol{\theta}} f(\textbf{w}_{k}) \right\|_2^2$ and $\mathbb{E} \left\| \nabla_{\boldsymbol{\theta}} f(\boldsymbol{\alpha}_{k},\boldsymbol{\theta}_{k-1}) \right\|_2^2$. We denote $2L^2 \rho^2 + 2\left\| \nabla_{\boldsymbol{\theta}} f(\boldsymbol{0})\right\|_2^2=m_1$ and $4L^2 \rho^2 + 2\left\| \nabla_{\boldsymbol{\theta}} f(\boldsymbol{0})\right\|_2^2=m_2$.\\

ASSUMPTION 3. We assume that for $k=1,2, \cdots$
 \begin{equation}
  \lambda_1 \boldsymbol{I} \preceq \boldsymbol{H}^{-1}_{k} \preceq \lambda_2 \boldsymbol{I}
 \end{equation}
 where $\boldsymbol{I}$ is the identity matrix and $\lambda_1, \lambda_2$ are positive scalars.
 The notation $\boldsymbol{A} \preceq \boldsymbol{B}$ means that $\boldsymbol{B}-\boldsymbol{A}$ is positive semidefinite.\\

ASSUMPTION 4. \textit{Define} $\boldsymbol\delta_{\boldsymbol{\alpha},k} = \nabla_{\boldsymbol{\alpha}} F(\boldsymbol{\alpha}_{k-1},\boldsymbol{\theta}_{k-1}, \xi_k) - \nabla_{\boldsymbol{\alpha}} f(\boldsymbol{\alpha}_{k-1},\boldsymbol{\theta}_{k-1})$ \textit{and} $\boldsymbol\delta_{\boldsymbol{\theta},k} = \nabla_{\boldsymbol{\theta}} F(\boldsymbol{\alpha}_{k},\boldsymbol{\theta}_{k-1}, \xi_k) - \nabla_{\boldsymbol{\theta}} f(\boldsymbol{\alpha}_{k},\boldsymbol{\theta}_{k-1})$. \textit{Here, $\xi_k, k=1,2,\cdots$ are independent identically distributed samples of realizations of $\xi$}. \textit{There exists a sequence {$\sigma_k$} and a constant $\varepsilon$ such that for any $k$}
\begin{align}
  \mathbb{E} \left\| \boldsymbol\delta_{\boldsymbol{\alpha},k} \right\|_2^2 \leq \sigma_k^2,~~ \mathbb{E} \left\| \boldsymbol\delta_{\boldsymbol{\theta},k} \right\|_2^2 \leq \sigma_k^2,\\
  \left\| \mathbb{E} \left[ \boldsymbol\delta_{\boldsymbol{\theta},k}| \boldsymbol\Xi_{k-1} \right] \right\|_2 \leq \varepsilon\gamma_k,  
\end{align}
where $\boldsymbol\Xi_{k-1}=\left(\xi_1,\xi_2,\cdots,\xi_{k-1}\right)$.

\textbf{Remark 4}. From Assumption 4 and Remark 1, we can obtain the boundedness of $\mathbb{E} \left\| \boldsymbol{g}_k \right\|_2^2$ from the following argument
 \begin{equation}
   \begin{aligned}
      \mathbb{E} \left\| \boldsymbol{g}_k \right\|_2^2 &= \left\| \nabla_{\boldsymbol{\alpha}} F(\boldsymbol{\alpha}_{k-1},\boldsymbol{\theta}_{k-1}, \xi_k) \right\|_2^2\\
     & = \mathbb{E} \left\| \boldsymbol\delta_{\boldsymbol{\alpha},k} + \nabla_{\boldsymbol{\alpha}} f(\boldsymbol{\alpha}_{k-1},\boldsymbol{\theta}_{k-1}) \right\|_2^2\\
     & \leq \mathbb{E} \left( 2 \left\| \boldsymbol\delta_{\boldsymbol{\alpha},k} \right\|_2^2 +2\left\| \nabla_{\boldsymbol{\alpha}} f(\boldsymbol{\alpha}_{k-1},\boldsymbol{\theta}_{k-1}) \right\|_2^2 \right)\\
     & \leq 2 \sigma_k^2 + 4L^2 \rho^2 + 4\left\| \nabla_{\boldsymbol{\alpha}} f(\boldsymbol{0})\right\|_2^2.
   \end{aligned}
 \end{equation}
 We denote $2 \sigma_k^2 + 4L^2 \rho^2 + 4\left\| \nabla_{\boldsymbol{\alpha}} f(\boldsymbol{0})\right\|_2^2=m_3$.\\
 
 Since the block updates (23) and (24) are Gauss-Seidel, the common assumption $\left\| \mathbb{E} \left[ \boldsymbol\delta_{\boldsymbol{\theta},k}| \boldsymbol\Xi_{k-1} \right] \right\|_2=0$ fails in our algorithm. However, the boundedness assumption (34) holds under proper condition \cite{ref44}. We give the case when $f(\textbf{w})=\frac{1}{M} \sum_{i=1}^{M}F(\textbf{w},\xi_i)$ with  $Prob\{\xi=\xi_i\}=\frac{1}{M}, i = 1,2, \cdots, M$. We also need that $F(\textbf{w},\xi_i)$ has Lipschitz continuous partial gradient 
 \begin{equation}
  \left\| \nabla_{\boldsymbol{\theta}} F(\textbf{w},\xi) - \nabla_{\boldsymbol{\theta}} F(\tilde{\textbf{w}},\xi) \right\|_2 \leq L\left\| \textbf{w}- \tilde{\textbf{w}} \right\|_2, \forall \textbf{w}, \tilde{\textbf{w}}.\notag
 \end{equation}
Then, we have
\begin{align}\notag
  & \mathbb{E} \left[ \nabla_{\boldsymbol{\theta}} F(\boldsymbol{\alpha}_{k},\boldsymbol{\theta}_{k-1}, \xi_k)|\boldsymbol\Xi_{k-1} \right]\\ \notag
   = & \sum_{i=1}^M Prob\{\xi_k=\xi_i\} \nabla_{\boldsymbol{\theta}} F(\boldsymbol{\alpha}_{k,i},\boldsymbol{\theta}_{k-1}, \xi_i)\\ \notag
   = & \frac{1}{M} \sum_{i=1}^M \nabla_{\boldsymbol{\theta}} F(\boldsymbol{\alpha}_{k,i},\boldsymbol{\theta}_{k-1}, \xi_i)
\end{align}
where $\boldsymbol{\alpha}_{k,i}=\boldsymbol{\alpha}_{k-1}-\gamma_k\boldsymbol{H}^{-1}_{k,i}\nabla_{\boldsymbol{\alpha}} F(\boldsymbol{\alpha}_{k-1},\boldsymbol{\theta}_{k-1}, \xi_i)$, and 
\begin{align}\notag
  & \mathbb{E} \left[ \nabla_{\boldsymbol{\theta}} f(\boldsymbol{\alpha}_{k},\boldsymbol{\theta}_{k-1})|\boldsymbol\Xi_{k-1} \right]\\ \notag
   = & \sum_{j=1}^M Prob\{\xi_k=\xi_j\} \nabla_{\boldsymbol{\theta}} f(\boldsymbol{\alpha}_{k,j},\boldsymbol{\theta}_{k-1})\\ \notag
   = & \frac{1}{M^2} \sum_{j=1}^M \sum_{i=1}^M \nabla_{\boldsymbol{\theta}} F(\boldsymbol{\alpha}_{k,j},\boldsymbol{\theta}_{k-1}, \xi_i)\\ \notag
   = & \frac{1}{M^2} \sum_{i=1}^M \sum_{j=1}^M \nabla_{\boldsymbol{\theta}} F(\boldsymbol{\alpha}_{k,j},\boldsymbol{\theta}_{k-1}, \xi_i)
\end{align}
Thus,
\begin{align}\notag
  &\left\| \mathbb{E} \left[ \boldsymbol\delta_{\boldsymbol{\theta},k}| \boldsymbol\Xi_{k-1} \right] \right\|_2\\\notag
   = & \left\| \mathbb{E} \left[ \nabla_{\boldsymbol{\theta}} F(\boldsymbol{\alpha}_{k},\boldsymbol{\theta}_{k-1}, \xi_k)- \nabla_{\boldsymbol{\theta}} f(\boldsymbol{\alpha}_{k},\boldsymbol{\theta}_{k-1})  | \boldsymbol\Xi_{k-1} \right] \right\|_2\\ \notag
   =  & \frac{1}{M^2} \left\| \sum_{i=1}^M \sum_{j=1}^M \left[ \nabla_{\boldsymbol{\theta}} F(\boldsymbol{\alpha}_{k,i},\boldsymbol{\theta}_{k-1}, \xi_i) - \nabla_{\boldsymbol{\theta}} F(\boldsymbol{\alpha}_{k,j},\boldsymbol{\theta}_{k-1}, \xi_i) \right] \right\|_2 \\ \notag
   \leq & \frac{L}{M^2} \sum_{i=1}^M \sum_{j=1}^M\left\| \boldsymbol{\alpha}_{k,i}- \boldsymbol{\alpha}_{k,j}\right\|_2 \\ \notag
   = & \frac{L}{M^2} \sum_{i=1}^M \sum_{j=1}^M \gamma_k \lVert \boldsymbol{H}^{-1}_{k,i} \nabla_{\boldsymbol{\alpha}} F(\boldsymbol{\alpha}_{k-1},\boldsymbol{\theta}_{k-1}, \xi_i) \\ \notag 
   & -\boldsymbol{H}^{-1}_{k,j}\nabla_{\boldsymbol{\alpha}} F(\boldsymbol{\alpha}_{k-1},\boldsymbol{\theta}_{k-1}, \xi_j)  \lVert_2 \\ \notag
   \leq & 2L\gamma_k \left\| \boldsymbol{H}^{-1}_{k}\nabla_{\boldsymbol{\alpha}} F(\boldsymbol{\alpha}_{k-1},\boldsymbol{\theta}_{k-1}, \xi_k) \right\|_2 
   \leq 2L\gamma_k \lambda_2 a.
 \end{align}
In the above we assume that $\left\| \nabla_{\boldsymbol{\alpha}} F(\boldsymbol{\alpha}_{k-1},\boldsymbol{\theta}_{k-1}, \xi_j) \right\|_2 \leq a $ for all $i,k$. Define $\varepsilon= 2L \lambda_2 a$, (34) is derived.

The following lemma, which can be found in \cite{ref45}, will be used in the proof of Theorem 1.\\

\textbf{Lemma 1.} Let $\{a_k\}$,$\{b_k\}$ be two non-negative real sequences, if $\sum_{k=1}^{+\infty} a_k=+\infty$, $\sum_{k=1}^{+\infty} a_k b_k < +\infty$, and there exists a constant $\kappa>0$ such that $|b_{k+1}-b_k|<\kappa a_k$, then 
\begin{align}
\lim_{k\rightarrow \infty} b_k=0. \notag
\end{align}\\

\textbf{Theorem 1.} \textit{Let $\{\textbf{w}_k\}$ be generated from (23) and (24), $\gamma_k=c_k\eta_k$, $\beta_k=d_k\eta_k$, and} 
\begin{align}
    &0<c<c_k<C, 0<d<d_k<D, \forall k;\\
	&\lim_{k \rightarrow \infty }=0, \sum_{k=1}^{+\infty} \eta_k = +\infty,~~ \sum_{k=1}^{+\infty} \eta^2_k < +\infty.
\end{align}
  
\textit{Under Assumptions 1 to 4, if $\sigma=\sup_k \sigma_k < \infty $, then }
\begin{equation}
  \lim_{k\rightarrow \infty} \mathbb{E} \left\| \nabla f(\textbf{w}_k) \right\|_2=0.
\end{equation}
\textit{Proof.} By the partial gradient Lipschitz continuity of $f$ in Assumption 1 and the descent lemma \cite{ref46}, we have 
\begin{align} \notag
  & f(\boldsymbol{\alpha}_{k},\boldsymbol{\theta}_{k-1}) - f(\boldsymbol{\alpha}_{k-1},\boldsymbol{\theta}_{k-1}) \\ \notag
   \leq & \langle \nabla_{\boldsymbol{\alpha}} f(\boldsymbol{\alpha}_{k-1},\boldsymbol{\theta}_{k-1}), -\gamma_k \boldsymbol{H}^{-1}_{k} \nabla_{\boldsymbol{\alpha}} F(\boldsymbol{\alpha}_{k-1},\boldsymbol{\theta}_{k-1}, \xi_k)\rangle\\ \notag
  & + \frac{L}{2}\left\| \boldsymbol{\alpha}_{k} - \boldsymbol{\alpha}_{k-1}\right\|_2^2 \\ \notag
  =& - \gamma_k \langle \nabla_{\boldsymbol{\alpha}} f(\boldsymbol{\alpha}_{k-1},\boldsymbol{\theta}_{k-1}), \boldsymbol{H}^{-1}_{k} \boldsymbol{g}_{k} \rangle  + \frac{L}{2}\left\| \gamma_k \boldsymbol{H}^{-1}_{k} \boldsymbol{g}_{k} \right\|_2^2\\ \notag
  & \leq - \gamma_k \langle \nabla_{\boldsymbol{\alpha}} f(\boldsymbol{\alpha}_{k-1},\boldsymbol{\theta}_{k-1}), \boldsymbol{H}^{-1}_{k} \left( \nabla_{\boldsymbol{\alpha}} f(\boldsymbol{\alpha}_{k-1},\boldsymbol{\theta}_{k-1}) + \boldsymbol\delta_{\boldsymbol{\alpha},k} \right) \rangle  \\ \notag
  & + \frac{L}{2} (\gamma_k \lambda_2)^2 \left\| \boldsymbol{g}_{k} \right\|_2^2 \\ \notag
  =& - \gamma_k \langle \nabla_{\boldsymbol{\alpha}} f(\boldsymbol{\alpha}_{k-1},\boldsymbol{\theta}_{k-1}), \boldsymbol{H}^{-1}_{k} \nabla_{\boldsymbol{\alpha}} f(\boldsymbol{\alpha}_{k-1},\boldsymbol{\theta}_{k-1}\rangle \\ \notag
  & - \gamma_k \langle \nabla_{\boldsymbol{\alpha}} f(\boldsymbol{\alpha}_{k-1},\boldsymbol{\theta}_{k-1}), \boldsymbol{H}^{-1}_{k} \boldsymbol\delta_{\boldsymbol{\alpha},k} \rangle + \frac{L}{2} (\gamma_k \lambda_2)^2 \left\| \boldsymbol{g}_{k} \right\|_2^2 \\ \notag
  \leq & -\gamma_k\lambda_1 \left\| \nabla_{\boldsymbol{\alpha}} f(\boldsymbol{\alpha}_{k-1},\boldsymbol{\theta}_{k-1}) \right\|_2^2 + \frac{L}{2} (\gamma_k \lambda_2)^2 \left\| \boldsymbol{g}_{k} \right\|_2^2\\ 
  & - \gamma_k \langle \nabla_{\boldsymbol{\alpha}} f(\boldsymbol{\alpha}_{k-1},\boldsymbol{\theta}_{k-1}), \boldsymbol{H}^{-1}_{k} \boldsymbol\delta_{\boldsymbol{\alpha},k} \rangle,
\end{align}
and
\begin{equation}
\begin{aligned}
  & f(\boldsymbol{\alpha}_{k},\boldsymbol{\theta}_{k}) - f(\boldsymbol{\alpha}_{k},\boldsymbol{\theta}_{k-1}) \\
  & \leq \langle \nabla_{\boldsymbol{\theta}} f(\boldsymbol{\alpha}_{k},\boldsymbol{\theta}_{k-1}), -\beta_k \nabla_{\boldsymbol{\theta}} F(\boldsymbol{\alpha}_k, \boldsymbol{\theta}_{k-1}, \xi_k) \rangle \\
  & + \frac{L}{2}\left\| \boldsymbol{\theta}_{k} - \boldsymbol{\theta}_{k-1}\right\|_2^2 \\
  & = - \beta_k \langle \nabla_{\boldsymbol{\theta}} f(\boldsymbol{\alpha}_{k},\boldsymbol{\theta}_{k-1}), \nabla_{\boldsymbol{\theta}} F(\boldsymbol{\alpha}_k, \boldsymbol{\theta}_{k-1}, \xi_k) \rangle  \\
  & + \frac{L}{2} \beta_k^2 \left\| \nabla_{\boldsymbol{\theta}} F(\boldsymbol{\alpha}_k, \boldsymbol{\theta}_{k-1}, \xi_k) \right\|_2^2\\
  & = - \beta_k \langle \nabla_{\boldsymbol{\theta}} f(\boldsymbol{\alpha}_{k},\boldsymbol{\theta}_{k-1}), \nabla_{\boldsymbol{\theta}} f(\boldsymbol{\alpha}_k, \boldsymbol{\theta}_{k-1})+ 
  \boldsymbol\delta_{\boldsymbol{\theta},k}\rangle  \\
  & + \frac{L}{2} \beta_k^2 \left\| \nabla_{\boldsymbol{\theta}} f(\boldsymbol{\alpha}_k, \boldsymbol{\theta}_{k-1}) + \boldsymbol\delta_{\boldsymbol{\theta},k} \right\|_2^2\\
  & \leq -(\beta_k - L \beta_k^2) \left\| \nabla_{\boldsymbol{\theta}} f(\boldsymbol{\alpha}_k, \boldsymbol{\theta}_{k-1}) \right\|_2^2 + L \beta_k^2 \left\| \boldsymbol\delta_{\boldsymbol{\theta},k} \right\|_2^2\\
  & - \beta_k \langle \nabla_{\boldsymbol{\theta}} f(\boldsymbol{\alpha}_k, \boldsymbol{\theta}_{k-1}), \boldsymbol\delta_{\boldsymbol{\theta},k} \rangle\\
  & =-(\beta_k - L \beta_k^2) \left\| \nabla_{\boldsymbol{\theta}} f(\boldsymbol{\alpha}_k, \boldsymbol{\theta}_{k-1}) \right\|_2^2 + L \beta_k^2 \left\| \boldsymbol\delta_{\boldsymbol{\theta},k} \right\|_2^2\\
  & - \beta_k \langle \nabla_{\boldsymbol{\theta}} f(\boldsymbol{\alpha}_k, \boldsymbol{\theta}_{k-1}) - \nabla_{\boldsymbol{\theta}} f(\boldsymbol{\alpha}_{k-1}, \boldsymbol{\theta}_{k-1}), \boldsymbol\delta_{\boldsymbol{\theta},k} \rangle\\
  & - \beta_k \langle \nabla_{\boldsymbol{\theta}} f(\boldsymbol{\alpha}_{k-1}, \boldsymbol{\theta}_{k-1}), \boldsymbol\delta_{\boldsymbol{\theta},k}\rangle\\
  & \leq -(\beta_k - L \beta_k^2) \left\| \nabla_{\boldsymbol{\theta}} f(\boldsymbol{\alpha}_k, \boldsymbol{\theta}_{k-1}) \right\|_2^2 \\
  &+ (L \beta_k^2 + \frac{L}{2} \beta_k\gamma_k \lambda_2) \left\| \boldsymbol\delta_{\boldsymbol{\theta},k} \right\|_2^2 + \frac{L}{2}\beta_k\gamma_k \lambda_2  \left\| \boldsymbol{g}_{k} \right\|_2^2 \\
  & - \beta_k  \langle \nabla_{\boldsymbol{\theta}} f(\boldsymbol{\alpha}_{k-1}, \boldsymbol{\theta}_{k-1}), \boldsymbol\delta_{\boldsymbol{\theta},k}\rangle.
\end{aligned}
\end{equation}
The last inequality in (40) follows from the fact that
\begin{equation}
  \begin{aligned}
  & - \beta_k \langle \nabla_{\boldsymbol{\theta}} f(\boldsymbol{\alpha}_k, \boldsymbol{\theta}_{k-1}) - \nabla_{\boldsymbol{\theta}} f(\boldsymbol{\alpha}_{k-1}, \boldsymbol{\theta}_{k-1}), \boldsymbol\delta_{\boldsymbol{\theta},k} \rangle\\
  & \leq \beta_k \left\| \nabla_{\boldsymbol{\theta}} f(\boldsymbol{\alpha}_k, \boldsymbol{\theta}_{k-1}) - \nabla_{\boldsymbol{\theta}} f(\boldsymbol{\alpha}_{k-1}, \boldsymbol{\theta}_{k-1}) \right\|_2 \left\| \boldsymbol\delta_{\boldsymbol{\theta},k} \right\|_2\\
  & \leq L \beta_k\left\| \boldsymbol\alpha_k- \boldsymbol\alpha_{k-1} \right\|_2 \left\| \boldsymbol\delta_{\boldsymbol{\theta},k} \right\|_2\\
  & =L\beta_k \left\| \gamma_k \boldsymbol{H}^{-1}_{k} \boldsymbol{g}_{k} \right\|_2 \left\| \boldsymbol\delta_{\boldsymbol{\theta},k} \right\|_2\\
  & \leq L\beta_k\gamma_k \lambda_2 \left\| \boldsymbol{g}_{k} \right\|_2 \left\| \boldsymbol\delta_{\boldsymbol{\theta},k} \right\|_2 \\
  & \leq \frac{L}{2} \beta_k\gamma_k \lambda_2 (\left\| \boldsymbol{g}_{k} \right\|_2^2 + \left\| \boldsymbol\delta_{\boldsymbol{\theta},k} \right\|_2^2).
  \end{aligned}\notag
\end{equation}

Summing (39) and (40), we get
\begin{align} \notag
  & f(\boldsymbol{\alpha}_{k},\boldsymbol{\theta}_{k}) - f(\boldsymbol{\alpha}_{k-1},\boldsymbol{\theta}_{k-1})\\ \notag
  & \leq -\gamma_k\lambda_1 \left\| \nabla_{\boldsymbol{\alpha}} f(\boldsymbol{\alpha}_{k-1},\boldsymbol{\theta}_{k-1}) \right\|_2^2 \\\notag
  & - \gamma_k \langle \nabla_{\boldsymbol{\alpha}} f(\boldsymbol{\alpha}_{k-1},\boldsymbol{\theta}_{k-1}), \boldsymbol{H}^{-1}_{k} \boldsymbol\delta_{\boldsymbol{\alpha},k} \rangle \\  \notag
  & - (\beta_k - L \beta_k^2) \left\| \nabla_{\boldsymbol{\theta}} f(\boldsymbol{\alpha}_k, \boldsymbol{\theta}_{k-1}) \right\|_2^2 \\ \notag
  & + (L \beta_k^2 + \frac{L}{2} \beta_k\gamma_k \lambda_2) \left\| \boldsymbol\delta_{\boldsymbol{\theta},k} \right\|_2^2 + \frac{L}{2}( \beta_k\gamma_k \lambda_2 +\gamma_k^2 \lambda_2^2)  \left\| \boldsymbol{g}_{k} \right\|_2^2 \\
  & - \beta_k  \langle \nabla_{\boldsymbol{\theta}} f(\boldsymbol{\alpha}_{k-1}, \boldsymbol{\theta}_{k-1}), \boldsymbol\delta_{\boldsymbol{\theta},k}\rangle.
 \end{align}

Note that
\begin{align} \notag
  & \mathbb{E} \langle \nabla_{\boldsymbol{\alpha}} f(\boldsymbol{\alpha}_{k-1},\boldsymbol{\theta}_{k-1}), \boldsymbol{H}^{-1}_{k} \boldsymbol\delta_{\boldsymbol{\alpha},k} \rangle\\ \notag
  & = \mathbb{E}_{\boldsymbol\Xi_{k-1}} \bigg[ \mathbb{E} \big[ \langle \nabla_{\boldsymbol{\alpha}} f(\boldsymbol{\alpha}_{k-1},\boldsymbol{\theta}_{k-1}), \boldsymbol{H}^{-1}_{k} \boldsymbol\delta_{\boldsymbol{\alpha},k} \rangle |\boldsymbol\Xi_{k-1} \big] \bigg]   \\\notag
  & = \mathbb{E}_{\boldsymbol\Xi_{k-1}} \bigg[ \langle \mathbb{E} \big[ \nabla_{\boldsymbol{\alpha}} f(\boldsymbol{\alpha}_{k-1},\boldsymbol{\theta}_{k-1})| \boldsymbol\Xi_{k-1} \big], \mathbb{E} \big[ \boldsymbol{H}^{-1}_{k} \boldsymbol\delta_{\boldsymbol{\alpha},k} | \boldsymbol\Xi_{k-1} \big] \rangle  \bigg] \\  \notag
  & \leq \mathbb{E}_{\boldsymbol\Xi_{k-1}} \bigg[ \left\| \mathbb{E} \big[ \nabla_{\boldsymbol{\alpha}} f(\boldsymbol{\alpha}_{k-1},\boldsymbol{\theta}_{k-1})| \boldsymbol\Xi_{k-1} \big] \right\|_2 \cdot\\ \notag
  & ~~~~~~~~~~~~\left\| \mathbb{E} \big[ \boldsymbol{H}^{-1}_{k} \boldsymbol\delta_{\boldsymbol{\alpha},k} | \boldsymbol\Xi_{k-1} \big] \right\| \bigg]\\ \notag
  & \leq \varepsilon\lambda_2\gamma_k \mathbb{E}_{\boldsymbol\Xi_{k-1}} \bigg[ \left\| \mathbb{E} \big[ \nabla_{\boldsymbol{\alpha}} f(\boldsymbol{\alpha}_{k-1},\boldsymbol{\theta}_{k-1})| \boldsymbol\Xi_{k-1} \big] \right\| \bigg]\\
  & \leq \varepsilon\lambda_2\gamma_k \mathbb{E} \left\| \nabla_{\boldsymbol{\alpha}} f(\boldsymbol{\alpha}_{k-1},\boldsymbol{\theta}_{k-1}) \right\|,
\end{align} 
 and
 \begin{align} \notag
  & \mathbb{E} \langle \nabla_{\boldsymbol{\theta}} f(\boldsymbol{\alpha}_{k-1},\boldsymbol{\theta}_{k-1}), \boldsymbol\delta_{\boldsymbol{\theta},k} \rangle\\ \notag
  & = \mathbb{E}_{\boldsymbol\Xi_{k-1}} \bigg[ \mathbb{E} \big[ \langle \nabla_{\boldsymbol{\theta}} f(\boldsymbol{\alpha}_{k-1},\boldsymbol{\theta}_{k-1}), \boldsymbol\delta_{\boldsymbol{\theta},k} \rangle |\boldsymbol\Xi_{k-1} \big] \bigg]   \\\notag
  & = \mathbb{E}_{\boldsymbol\Xi_{k-1}} \bigg[ \langle \mathbb{E} \big[ \nabla_{\boldsymbol{\theta}} f(\boldsymbol{\alpha}_{k-1},\boldsymbol{\theta}_{k-1})| \boldsymbol\Xi_{k-1} \big], \mathbb{E} \big[ \boldsymbol\delta_{\boldsymbol{\theta},k} | \boldsymbol\Xi_{k-1} \big] \rangle  \bigg] \\  \notag
  & \leq \mathbb{E}_{\boldsymbol\Xi_{k-1}} \bigg[ \left\| \mathbb{E} \big[ \nabla_{\boldsymbol{\theta}} f(\boldsymbol{\alpha}_{k-1},\boldsymbol{\theta}_{k-1})| \boldsymbol\Xi_{k-1} \big] \right\| \cdot \left\| \mathbb{E} \big[ \boldsymbol\delta_{\boldsymbol{\theta},k} | \boldsymbol\Xi_{k-1} \big] \right\| \bigg]\\ \notag
  & \leq \varepsilon\gamma_k \mathbb{E}_{\boldsymbol\Xi_{k-1}} \bigg[ \left\| \mathbb{E} \big[ \nabla_{\boldsymbol{\theta}} f(\boldsymbol{\alpha}_{k-1},\boldsymbol{\theta}_{k-1})| \boldsymbol\Xi_{k-1} \big] \right\| \bigg]\\
  & \leq \varepsilon\gamma_k \mathbb{E} \left\| \nabla_{\boldsymbol{\theta}} f(\boldsymbol{\alpha}_{k-1},\boldsymbol{\theta}_{k-1}) \right\|.
\end{align} 

Taking expectation on (41) and using (42) and (43), we have
 \begin{align} \notag
  & \mathbb{E} f(\textbf{w}_{k}) - \mathbb{E} f(\textbf{w}_{k-1})\\ \notag
  \leq & -\gamma_k\lambda_1 \mathbb{E}\left\| \nabla_{\boldsymbol{\alpha}} f(\boldsymbol{\alpha}_{k-1},\boldsymbol{\theta}_{k-1}) \right\|_2^2 \\\notag
  & + \gamma_k^2\varepsilon\lambda_2 \mathbb{E} \left\| \nabla_{\boldsymbol{\alpha}} f(\boldsymbol{\alpha}_{k-1},\boldsymbol{\theta}_{k-1}) \right\|_2 \\  \notag
  & - (\beta_k - L \beta_k^2) \mathbb{E} \left\| \nabla_{\boldsymbol{\theta}} f(\boldsymbol{\alpha}_k, \boldsymbol{\theta}_{k-1}) \right\|_2^2 \\ \notag
  & + (L \beta_k^2 + \frac{L}{2} \beta_k\gamma_k \lambda_2) \mathbb{E}  \left\| \boldsymbol\delta_{\boldsymbol{\theta},k} \right\|_2^2 \\ \notag
  & + \frac{L}{2}( \beta_k\gamma_k \lambda_2 +\gamma_k^2 \lambda_2^2) \mathbb{E} \left\| \boldsymbol{g}_{k} \right\|_2^2 \\ \notag
  & - \beta_k\gamma_k\varepsilon \mathbb{E} \left\| \nabla_{\boldsymbol{\theta}} f(\boldsymbol{\alpha}_{k-1},\boldsymbol{\theta}_{k-1}) \right\|\\ \notag
   \leq & -c\eta_k\lambda_1 \mathbb{E}\left\| \nabla_{\boldsymbol{\alpha}} f(\boldsymbol{\alpha}_{k-1},\boldsymbol{\theta}_{k-1}) \right\|_2^2 \\\notag
  & + C^2\eta_k^2\varepsilon\lambda_2 \mathbb{E} \left\| \nabla_{\boldsymbol{\alpha}} f(\boldsymbol{\alpha}_{k-1},\boldsymbol{\theta}_{k-1}) \right\|_2 \\  \notag
  & - (d\eta_k - LD^2 \eta_k^2) \mathbb{E} \left\| \nabla_{\boldsymbol{\theta}} f(\boldsymbol{\alpha}_k, \boldsymbol{\theta}_{k-1}) \right\|_2^2 \\ \notag
  & + (L D^2\eta_k^2 + \frac{L}{2} CD\eta_k^2 \lambda_2) \mathbb{E}  \left\| \boldsymbol\delta_{\boldsymbol{\theta},k} \right\|_2^2 \\ \notag
  & + \frac{L}{2}( CD\eta_k^2 \lambda_2 + C^2\eta_k^2 \lambda_2^2) \mathbb{E} \left\| \boldsymbol{g}_{k} \right\|_2^2 \\
  & - cd\eta_k^2\varepsilon \mathbb{E} \left\| \nabla_{\boldsymbol{\theta}} f(\boldsymbol{\alpha}_{k-1},\boldsymbol{\theta}_{k-1}) \right\|
 \end{align}

Summing (44) over $k$ and using (28), (29), (31), (33), (35), (36), (37) and that $f$ is lower bounded, we have
\begin{align}
  \sum_{k=1}^\infty \eta_k \mathbb{E} \left\| \nabla_{\boldsymbol{\alpha}} f(\boldsymbol{\alpha}_{k-1},\boldsymbol{\theta}_{k-1}) \right\|_2^2<\infty, \\
  \sum_{k=1}^\infty \eta_k \mathbb{E} \left\| \nabla_{\boldsymbol{\theta}} f(\boldsymbol{\alpha}_{k},\boldsymbol{\theta}_{k-1}) \right\|_2^2<\infty.
 \end{align}
 
Moreover,
\begin{align} \notag
  & \left | \mathbb{E} \left\| \nabla_{\boldsymbol{\alpha}} f(\boldsymbol{\alpha}_{k},\boldsymbol{\theta}_{k}) \right\|_2^2 - \mathbb{E} \left\| \nabla_{\boldsymbol{\alpha}} f(\boldsymbol{\alpha}_{k-1},\boldsymbol{\theta}_{k-1}) \right\|_2^2\right | \\ \notag 
  & =  |  \mathbb{E} \langle \nabla_{\boldsymbol{\alpha}} f(\boldsymbol{\alpha}_{k},\boldsymbol{\theta}_{k}) + \nabla_{\boldsymbol{\alpha}} f(\boldsymbol{\alpha}_{k-1},\boldsymbol{\theta}_{k-1}) \\ \notag 
  & \nabla_{\boldsymbol{\alpha}} f(\boldsymbol{\alpha}_{k},\boldsymbol{\theta}_{k}) - \nabla_{\boldsymbol{\alpha}} f(\boldsymbol{\alpha}_{k-1},\boldsymbol{\theta}_{k-1})\rangle | \\\notag
  & \leq  \mathbb{E} \big[ \left\| \nabla_{\boldsymbol{\alpha}} f(\boldsymbol{\alpha}_{k},\boldsymbol{\theta}_{k}) + \nabla_{\boldsymbol{\alpha}} f(\boldsymbol{\alpha}_{k-1},\boldsymbol{\theta}_{k-1}) \right\|_2 \cdot \\  \notag
  &  \left\| \nabla_{\boldsymbol{\alpha}} f(\boldsymbol{\alpha}_{k},\boldsymbol{\theta}_{k}) - \nabla_{\boldsymbol{\alpha}} f(\boldsymbol{\alpha}_{k-1},\boldsymbol{\theta}_{k-1}) \right\|_2 \big] \\ \notag
  & \leq 2L\sqrt{m_1} \mathbb{E} \left\| \boldsymbol{\alpha}_{k} - \boldsymbol{\alpha}_{k-1} \right\|_2 \\ \notag
  & = 2L\sqrt{m_1} \mathbb{E} \left\| \gamma_k \boldsymbol{H}^{-1}_{k} \boldsymbol{g}_{k} \right\|_2 \\
  & \leq 2LC\eta_k\sqrt{m_1m_3}\lambda_2,
 \end{align}
 and similarily
 \begin{align} \notag
  & \left | \mathbb{E} \left\| \nabla_{\boldsymbol{\theta}} f(\boldsymbol{\alpha}_{k+1},\boldsymbol{\theta}_{k}) \right\|_2^2 - \mathbb{E} \left\| \nabla_{\boldsymbol{\theta}} f(\boldsymbol{\alpha}_{k},\boldsymbol{\theta}_{k-1}) \right\|_2^2\right | \\ \notag 
   \leq & 2L\eta_k\sqrt{m_2}\sqrt{C^2\lambda^2m_3+2D^2(m_2+\sigma_k^2)}.
 \end{align}
 By Lemma 1, (45), (46), (47), and (48) yields 
 \begin{align}
  \lim_{k\rightarrow \infty} \mathbb{E} \left\| \nabla_{\boldsymbol{\alpha}} f(\boldsymbol{\alpha}_{k},\boldsymbol{\theta}_{k}) \right\|_2^2 = 0 \\ 
  \lim_{k\rightarrow \infty} \mathbb{E} \left\| \nabla_{\boldsymbol{\theta}} f(\boldsymbol{\alpha}_{k},\boldsymbol{\theta}_{k-1}) \right\|_2^2 = 0.
 \end{align}
 By Jensen's inequality, (48) implies
 \begin{align}
  \lim_{k\rightarrow \infty} \mathbb{E} \left\| \nabla_{\boldsymbol{\alpha}} f(\boldsymbol{\alpha}_{k},\boldsymbol{\theta}_{k}) \right\|_2 = 0. 
 \end{align}
 And
 \begin{align} \notag
  & \mathbb{E}  \left\| \nabla_{\boldsymbol{\theta}} f(\boldsymbol{\alpha}_{k},\boldsymbol{\theta}_{k}) \right\|_2 \\
  \leq & \mathbb{E}  \left\| \nabla_{\boldsymbol{\theta}} f(\boldsymbol{\alpha}_{k},\boldsymbol{\theta}_{k})- \nabla_{\boldsymbol{\theta}} f(\boldsymbol{\alpha}_{k},\boldsymbol{\theta}_{k-1}) \right\|_2 \\
   &+ \mathbb{E} \left\| \nabla_{\boldsymbol{\theta}} f(\boldsymbol{\alpha}_{k},\boldsymbol{\theta}_{k-1}) \right\|_2 \\ \notag 
  \leq & L \left\| \boldsymbol{\theta}_{k} - \boldsymbol{\theta}_{k-1} \right\|_2 + \mathbb{E} \left\| \nabla_{\boldsymbol{\theta}} f(\boldsymbol{\alpha}_{k},\boldsymbol{\theta}_{k-1}) \right\|_2 \\
  \leq & LD \eta_k\sqrt{2(m_2+\sigma_k^2)} + \mathbb{E} \left\| \nabla_{\boldsymbol{\theta}} f(\boldsymbol{\alpha}_{k},\boldsymbol{\theta}_{k-1}) \right\|_2
 \end{align}
 implies
 \begin{align}
  \lim_{k\rightarrow \infty} \mathbb{E} \left\| \nabla_{\boldsymbol{\theta}} f(\boldsymbol{\alpha}_{k},\boldsymbol{\theta}_{k}) \right\|_2 = 0. 
 \end{align}
 Therefore, the proof of Theorem 1 has been completed. $\hfill\blacksquare$
 
\subsection{Convergence Rate and Complexity}
Since we assume $f$ is nonconvex, we only discuss the convergence speed of final phase of the process. Therefore, we assume that $\textbf{w}_k$ are confined in a bounded domain where $f(\textbf{w})$ has a single minimum. 
It is well-known that classic SGD exhibits an asymptotically optimal convergence rate $O(1/k)$ in terms of expected objective value under sufficient regularity conditions and the second-order SGD only improves the constant \cite{ref20, ref46, ref47}. Thus, the proposed Algorithm 2 is expected to behave like $O(1/k)$. For updating $\boldsymbol{\alpha}$, our proposed Algorithm 2 requires $O(p^2)$ space and time per iteration; for $\boldsymbol{\theta}$, $O(q)$ space and time per iteration is required. Usually, $p$ in the machine learning tasks is relatively small, and thus the computational load is affordable. 

The algorithm may not scale well with large $p$. However, we can overcome this limitation by using stochastic quasi-Newton methods \cite{ref48}, such as online limited memory BFGS algorithms \cite{ref2, ref36, ref39, ref49}, which requires a complexity of $O(lp)$ in space and time. Regarding improving the convergence rate of the algorithm, we may consider implementing variance reduction techniques such as those described in \cite{ref50, ref51} in the future.

\section{Experiments}
In this section, we present empirical results on two classic tasks, regression and classification, to demonstrate the efficiency and effectiveness of our proposed SepSA algorithm. We compared it with widely-used neural network training algorithms, including the ordinary stochastic gradient descent method without momentum (SGD), Nesterov accelerated gradient (NAG) \cite{ref52} which computes the gradients at the predicted point instead of the current point, as well as two adaptive learning rate algorithms: RMSprop \cite{53}, which divides the learning rate by an exponentially decaying average of squared gradients, and Adaptive Moment Estimation (Adam) \cite{ref54}. In addition, we also compare our algorithm with the recently proposed slimTrain \cite{ref32} algorithm which also utilizes network separability. The experimental results demonstrate the strong competitiveness of the SepSA algorithm as it exhibits notable advantages such as faster convergence speed, less sensitivity to learning rate, and greater suitability for online learning.

\subsection{Data sets}
We selected four datasets from the PyTorch dataset library: the regression dataset Energy efficiency (referred to as Energy for simplicity) and Diabetes, and the classification datasets MNIST and CIFAR-10. The Energy dataset \cite{ref55} comprises 8 features, such as relative humidity and ambient temperature, and contains two regression targets: Cooling Load and Heating Load. The Diabetes dataset \cite{ref56}, provided by the National Institute of Diabetes and Digestive and Kidney Diseases (NIDDK), includes 10 features such as age, sex, BMI, and various other biomedical indicators, which measure the disease progression of patients through numerical labels. The MNIST dataset \cite{ref57} is a widely-used computer vision dataset comprising grayscale images depicting handwritten digits. The dataset contains 60,000 training images of 28$\times$28 pixels, and 10,000 testing images, each labeled with a corresponding number from 0-9. The CIFAR-10 dataset \cite{ref58} is frequently utilized to perform object recognition tasks in computer vision. The dataset includes 50,000 color images of 32$\times$32 pixels as a training set and 10,000 testing images. The images are categorized into 10 different classes, with each comprising 5,000 training images and 1,000 testing images. Table I summarizes the dimensions of the input and output of these datasets, along with the number of samples used during training and testing in our experiments.
  
\begin{table}[h]
	\centering
	\renewcommand\arraystretch{1.5}
	\caption{}
	\begin{tabular}{|c|c|c|c|c|c|}\hline
		Dataset & \makecell[c]{Type of\\Task} & \makecell[c]{Input\\Dim}  & \makecell[c]{Output\\Dim}& \makecell[c]{Training\\Samples} & \makecell[c]{Testing\\Samples} \\ \hline
		Energy & Regression& 8& 2& 491& 154 \\
		Diabetes & Regression&	10&	1 & 282 & 89 \\
		MNIST	& Classification&	28$\times$28& 10& 60,000& 10,000\\
		CIFAR-10 & Classification&	32$\times$32& 10& 50,000& 10,000\\
		\hline
	\end{tabular}
\end{table}

\subsection{Experimental Setup}
 For the regression and classification datasets, we employed feed-forward neural networks (FNN) and convolutional neural networks (CNN), respectively. The network structure is shown in Table II. Since the regression dataset has a small sample size, we used a single hidden layer FNN with 50 neurons and ReLU activation function. The output layer of the FNN uses linear neurons. The CNN comprises two convolutional layers, two pooling layers, and two fully connected layers. Batch normalization is applied to the first and third layers to accelerate the training process. The first convolutional layer utilizes a kernel size of 3$\times$3 with stride 1, padding 1, and 32 output channels to extract features from the input data. This is followed by batch normalization and ReLU activation function. To reduce the dimensionality of feature maps, we implemented a max pooling layer with a window size of 2$\times$2. The subsequent convolutional layer employs 64 output channels and the same convolutional and activation settings as the previous layer. The output is then processed through a fully connected layer, which flattens the pooled feature maps and connects them to 128 hidden nodes using ReLU activation. Lastly, the output layer uses an affine mapping and has 10 nodes to classify the output.

It is well-known that tedious hyper-parameter tuning is necessary to achieve satisfactory performance when training a neural network model. Different tasks may require different optimal hyper-parameters for each algorithm. Therefore, in our experiments, we evaluated the performance of various algorithms using multiple fixed learning rates (i.e., $lr=1e-2$, $1e-3$, and $1e-4$). We set other hyper-parameters of the algorithms to common default values. For example, both NAG and RMSprop used a momentum coefficient of 0.9, while Adam used two momentum coefficients of 0.9 and 0.999 for $\beta_1$ and $\beta_2$, respectively. We typically set a memory depth of $r=5$ for the slimTrain algorithm, which is consistent with its experimental settings in \cite{ref32}. However, during our experiments with online learning on MNIST and CIFAR-10 datasets, we have to set $r=0$ for \texttt{slimTrain} due to singular problems in the SVD decomposition. We adopt the mean square error loss as the objective function, and initialize the network weight using the Kaiming uniform method \cite{ref59}. We ensure the same initial conditions by setting a common random seed. For more details on the algorithm settings and code implementation, kindly visit our GitHub page.

\begin{table*}[b]
	\centering
	\caption{Structure of networks used in experiments. $d_i$ and $d_o$ indicates the input and output dimensions of the networks, and $C_i$ denotes the input channels of CNN.}
	\begin{tabular}{|c|c|c|c|}\hline
		Network Type &Layer Type & Description & Output Features \\ \hline
		\hline
		\multirow{2}{*}{FNN} & Affine + ReLU &$50\times d_i$matrix + $50\times 1$ bias& 50 \\ \cline{2-4}
		& Affine &$d_o\times 50$ matrix + $d_o\times 1$ bias& $d_o$ \\ \hline
		\hline
		\multirow{6}{*}{CNN}
		&Conv. + BatchNorm + ReLU&32, 3$\times$3$\times C_i$ filters, stride 1 padding 1&$d_i\times d_i\times 32$\\ \cline{2-4}
		&Max Pool&2$\times$2 pool&$d_i/2\times d_i/2\times 32$\\ \cline{2-4}
		&Conv. + BatchNorm + ReLU&64, 3$\times$3$\times$32 filters, stride 1 padding 1&$d_i/2\times d_i/2\times 64$\\ \cline{2-4}
		&Max Pool&2$\times$2 pool&$d_i/4\times d_i/4\times 64$\\ \cline{2-4}
		&Affine + ReLU&128$\times$($d_i/4\times d_i/4\times 64$) matrix + 128$\times$1 bias&128\\ \cline{2-4}
		&Affine&10$\times$128 matrix + 10$\times$1 bias&10\\ \hline
	\end{tabular}
\end{table*}

\subsection{Results of online learning (one sample at each iteration)}
Fig. 1 illustrates the online learning results of the algorithms on four different datasets with different learning rate of $1e-2, 1e-3$ and $1e-4$. The figure presents the test results (mean squared error, MSE) of each iteration. It is clear from the figure that the SepSA algorithm outperforms other algorithms on all datasets and at different learning rates. It exhibits the fastest convergence speed and is the least sensitive to the learning rate. In contrast, the \texttt{slimTrain} algorithm also shows some insensitivity to the learning rate, but its performance is not as good as that of SepSA and displays significant oscillations, primarily due to the use of small batch sizes \cite{ref32}. When batch size is small, slimTrain's approach of solving the approximate optimal solution of the parameters of the last layer for every iteration is more likely to cause severe oscillations. Despite the significant improvement in performance during mini-batch learning (Section IV-D), slimTrain's relatively poor adaptability to online learning is apparent.

\begin{figure*}[htp]
	\centering
	\includegraphics[width=1\linewidth, height=0.25\textwidth]{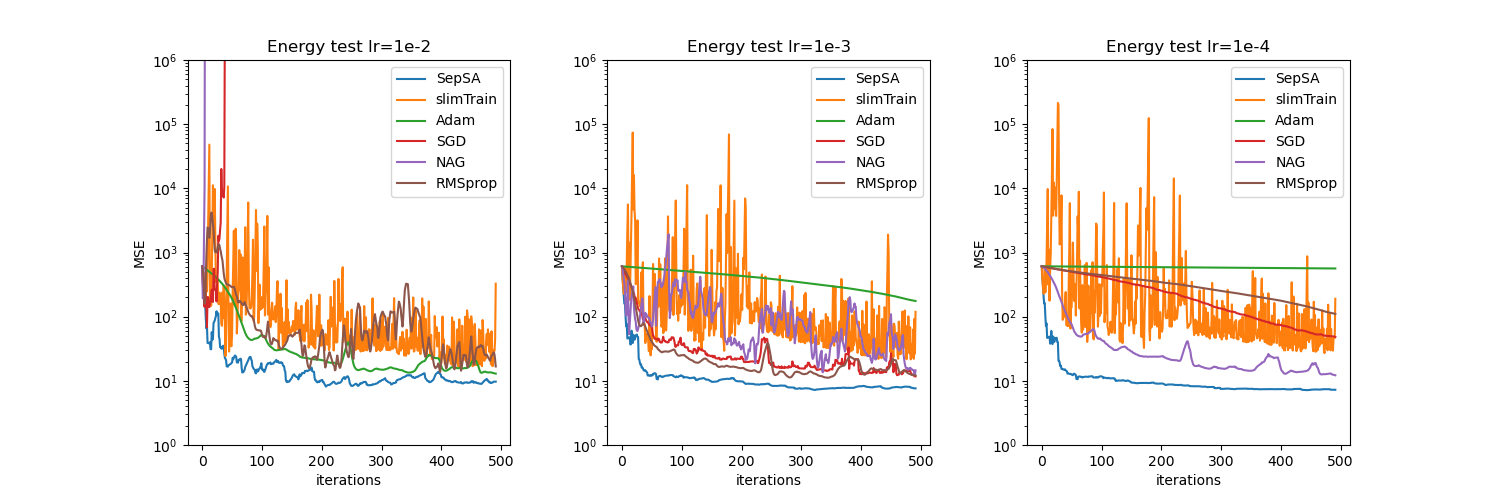}
	\includegraphics[width=1\linewidth, height=0.25\textwidth]{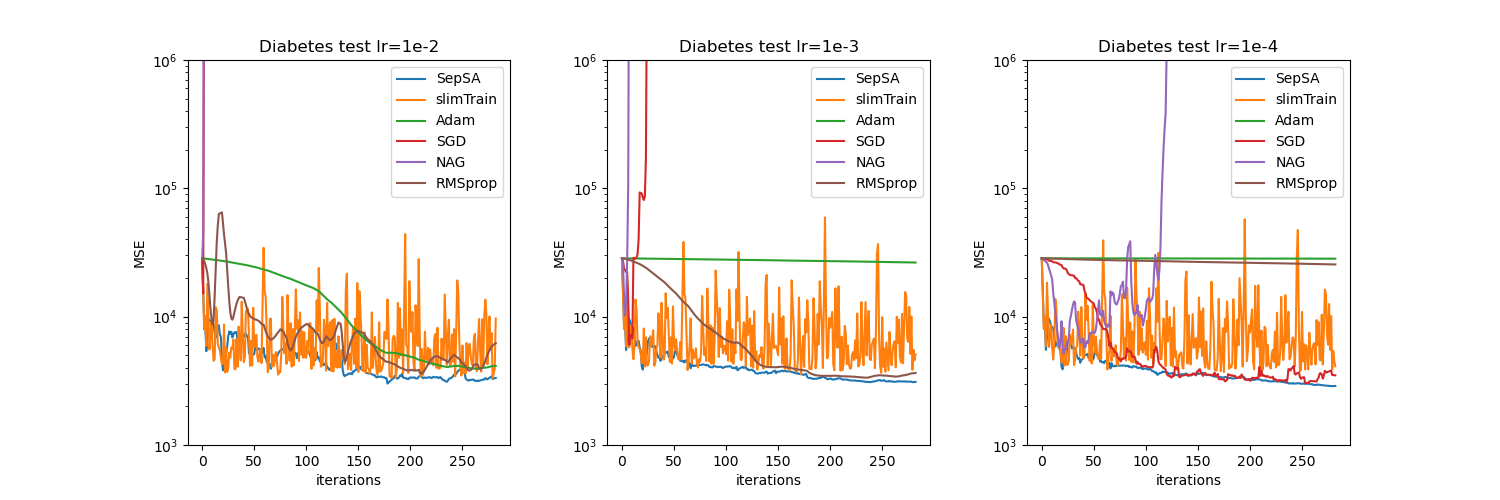}
	\includegraphics[width=1\linewidth, height=0.25\textwidth]{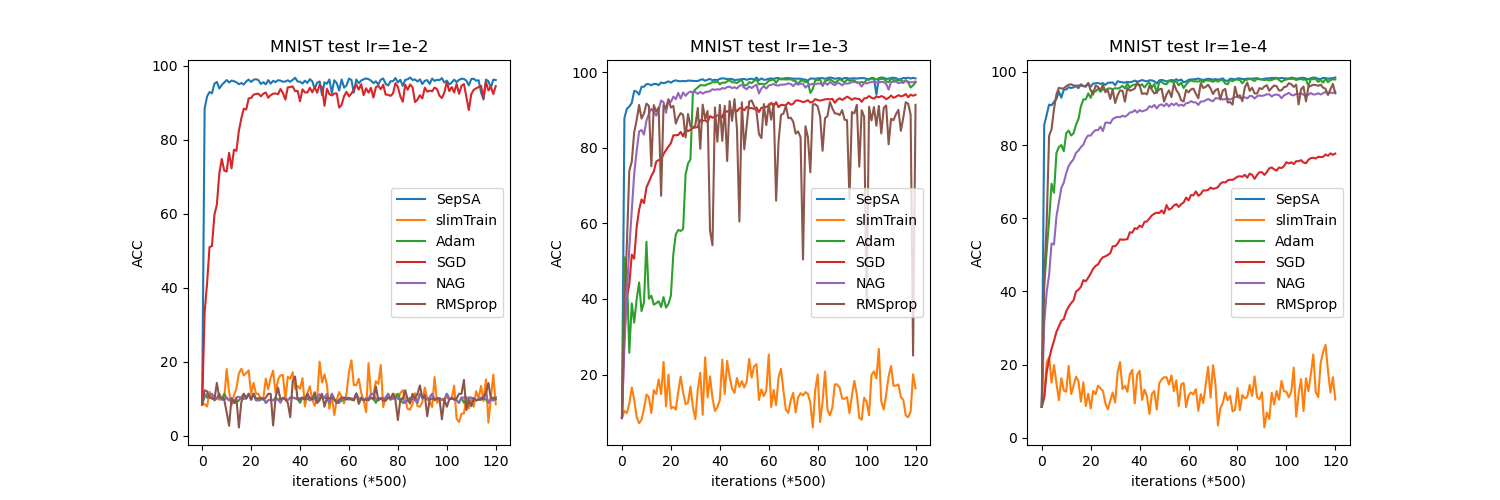}
	\includegraphics[width=1\linewidth, height=0.25\textwidth]{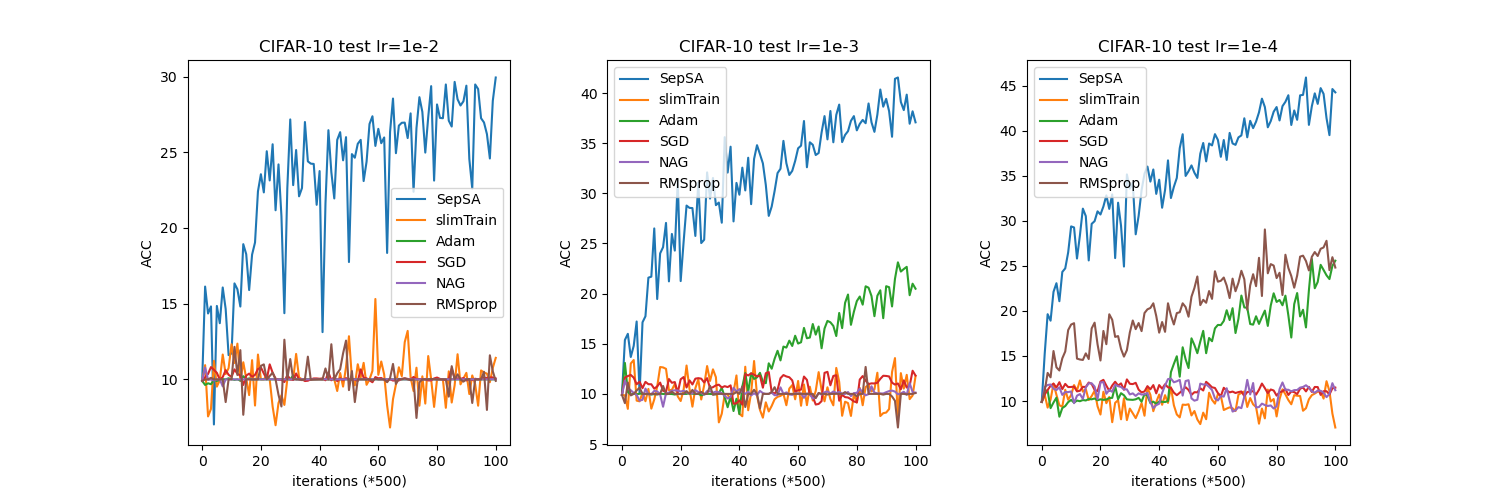}
	\caption{Online learning results of various algorithms on four datasets, each with different learning rates ($lr$). Due to the large sample sizes of MNIST and CIFAR-10, we plot a point every 500 iterations, and ACC represents classification prediction accuracy.}
\end{figure*}

The performance of other algorithms (Adam, SGD, NAG, RMSprop) varies across different datasets and learning rates, and it highly depends on the learning rate, particularly for non-adaptive learning rate methods like SGD and NAG. For instance, in the Energy dataset, when $lr=1e-2$, Adam exhibits the second-best performance, while SGD and NAG diverge due to the learning rate being too high for them. When $lr=1e-3$, RMSprop and SGD perform well, and when the learning rate is reduced further to $1e-4$, NAG exhibits better results due to the acceleration of Nesterov momentum.

For the Diabetes dataset, we found that the appropriate learning rate for the Adam algorithm is $1e-2$ or even higher, whereas RMSprop and SGD perform best with learning rates of $1e-3$ and $1e-4$, respectively. The NAG algorithm requires a smaller learning rate as it diverges under the above-three learning rate settings. In the MNIST and CIFAR-10 datasets, we observed that if the learning rate is too high, algorithms that are sensitive to the learning rate are susceptible to divergence, resulting in maintaining the test accuracy at around 10\% with oscillation. Nearly all algorithms performed poorly on CIFAR-10, and only our SepSA algorithm showed superior performance to other algorithms due to its learning rate insensitivity and faster convergence speed.

\subsection{Results of mini-batch learning}
To provide a comprehensive evaluation of the algorithms, we present the mini-batch learning results with multiple epochs using various learning rates. Note that SepSA has lower time complexity in comparison to slimTrain which requires SVD operations. However, the batch processing increases computational burden due to the RLS's recursive processing of multiple samples. To improve the efficiency of SepSA for mini-batch learning, we implement an approach where the batch size exponentially decays when updating $\boldsymbol{\alpha}$. Specifically, during the $i$-th epoch training process, for each iteration, only  $\lceil 0.5^{(i-1)}*\text{(batch size)} \rceil$ randomly selected samples are used for RLS.

Fig. 2 shows the mini-batch learning results on the regression datasets with different learning rates. It can be observed that in most cases, SepSA outperforms the other algorithms in terms of accuracy and convergence speed. The second-best algorithm is slimTrain whose performance has considerably improved compared to online learning after the batch size has much increased. Both SepSA and slimTrain have faster initial convergence speeds than the other methods. Similar to online learning, the performance of other algorithms highly depends on the appropriate learning rate. Even with a proper learning rate, their convergence speed is slower compared to SepSA. Note that the algorithms trained on the Diabetes dataset are slightly overfitting, which can be resolved by adjusting the regularization parameters, but it is not the focus of this article. Additionally, the results of some algorithms with the learning rate of $1e-2$ are not shown because the step size is too large, causing them to diverge. 

\begin{figure*}[htp]
	\centering
	\noindent\colorbox{gray!20}{\makebox[\dimexpr0.81\linewidth-2\fboxsep-2\fboxrule]{Energy train}}
	\includegraphics[width=1\linewidth]{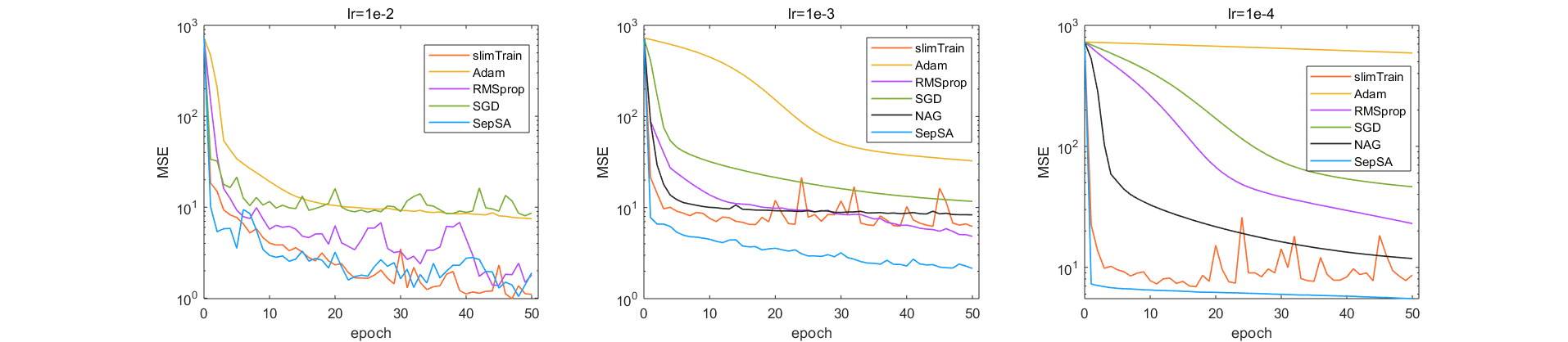}
	\noindent\colorbox{gray!20}{\makebox[\dimexpr0.81\linewidth-2\fboxsep-2\fboxrule]{Energy test}}
	\includegraphics[width=1\linewidth]{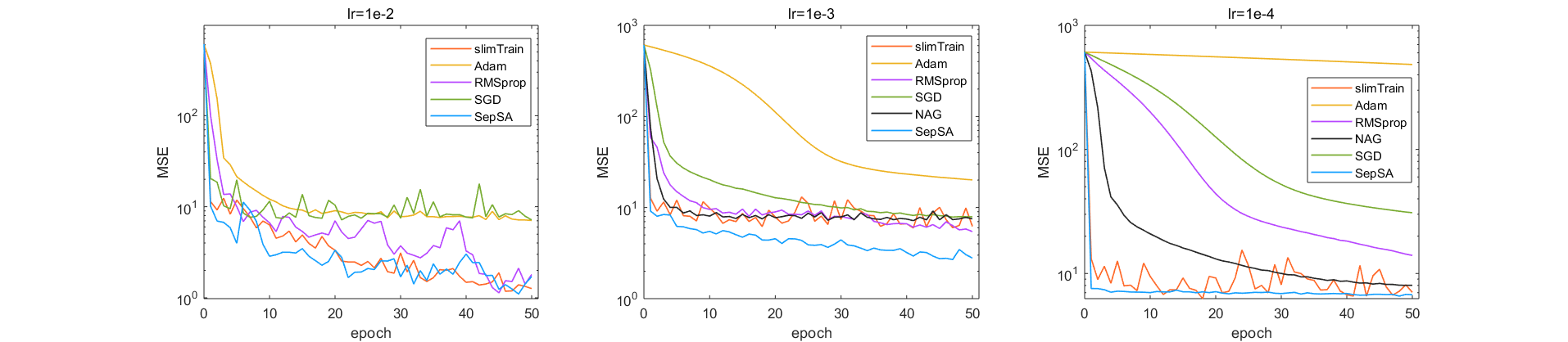}
	\noindent\colorbox{gray!20}{\makebox[\dimexpr0.81\linewidth-2\fboxsep-2\fboxrule]{Diabetes train}}
	\includegraphics[width=1\linewidth]{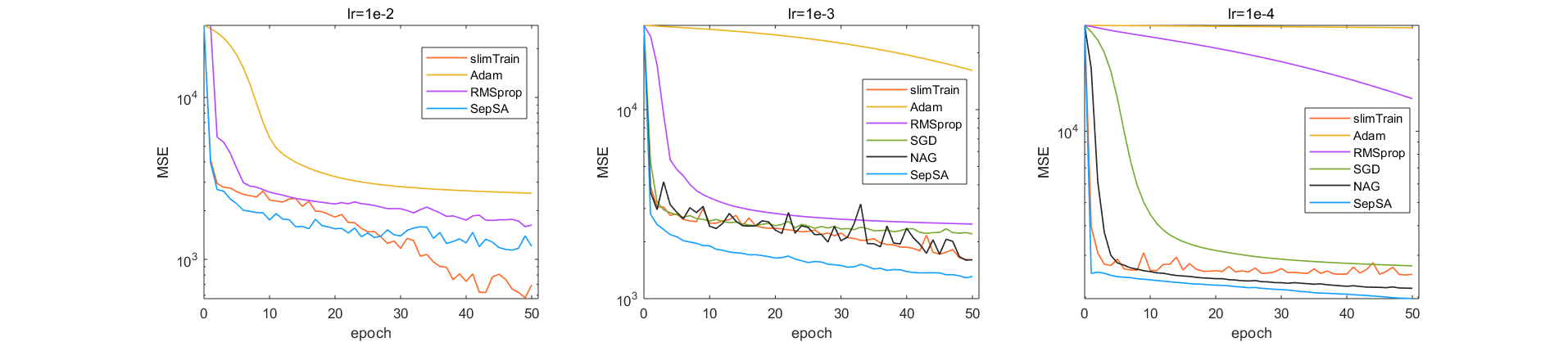}
	\noindent\colorbox{gray!20}{\makebox[\dimexpr0.81\linewidth-2\fboxsep-2\fboxrule]{Diabetes test}}
	\includegraphics[width=1\linewidth]{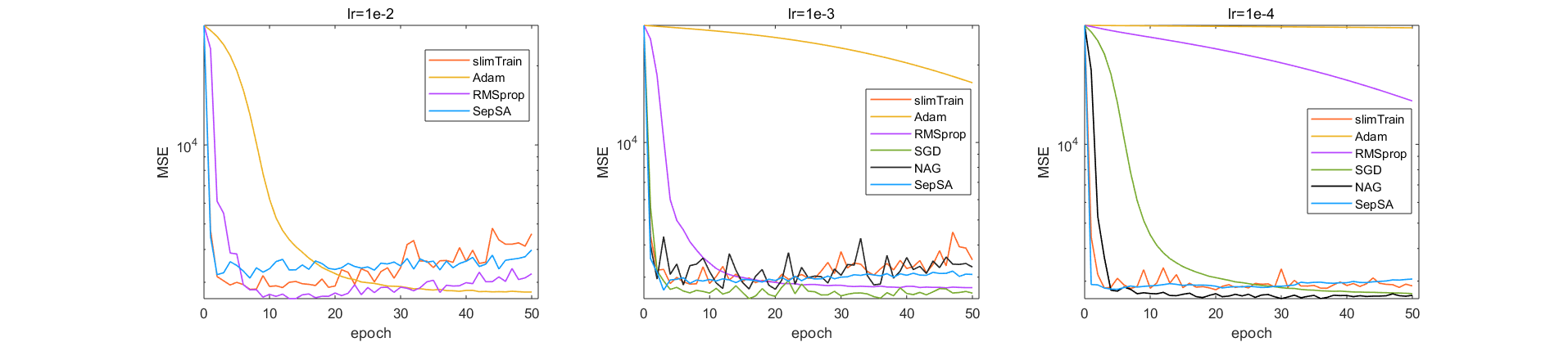}
	\caption{Mini-batch learning results of various algorithms on the regression datasets with different learning rates. The batch size is 32.}
\end{figure*}

The results on the classification datasets demonstrate a similar trend, as shown in Fig. 3. SepSA achieves significantly higher classification accuracy than the other algorithms at the end of the first epoch, and it maintains excellent training performance even with different learning rates. In contrast, the performance of other algorithms vary under different conditions. It is worth noting that the effect of slimTrain has deteriorated compared to the results on the regression dataset, and its initial convergence rate has slowed down. This may be due to the poor condition number of the matrix decomposed by slimTrain when solving $\hat{\boldsymbol{W}}(\boldsymbol{\theta})$ using SVD.

\begin{figure*}[htp]
	\centering
	\noindent\colorbox{gray!20}{\makebox[\dimexpr0.81\linewidth-2\fboxsep-2\fboxrule]{MNIST train}}
	\includegraphics[width=1\linewidth]{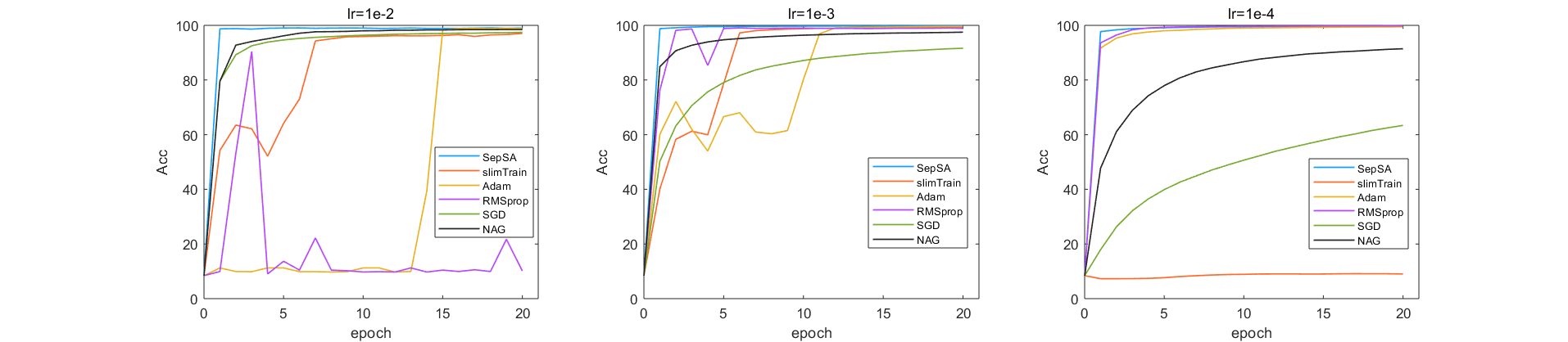}
	\noindent\colorbox{gray!20}{\makebox[\dimexpr0.81\linewidth-2\fboxsep-2\fboxrule]{MNIST test}}
	\includegraphics[width=1\linewidth]{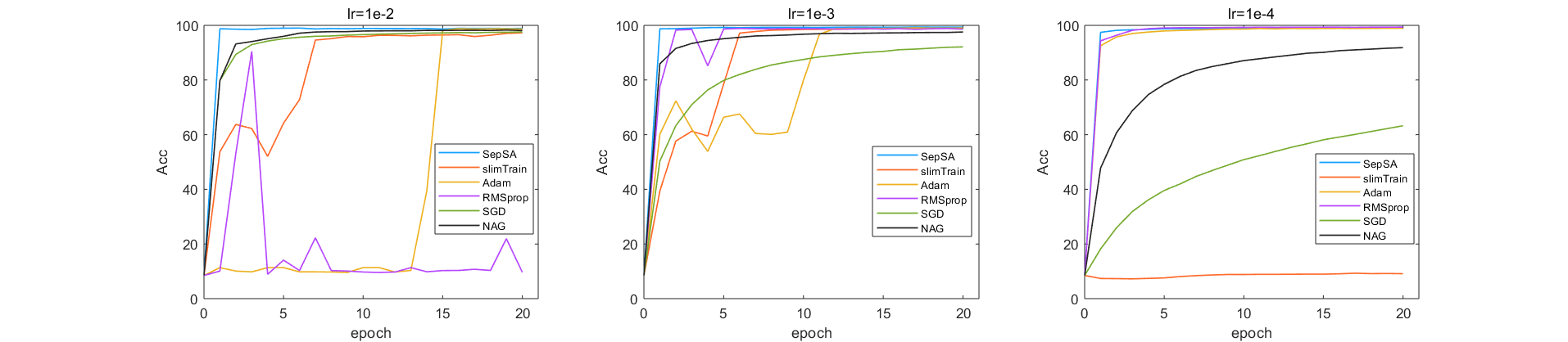}
	\noindent\colorbox{gray!20}{\makebox[\dimexpr0.81\linewidth-2\fboxsep-2\fboxrule]{CIFAR-10 train}}
	\includegraphics[width=1\linewidth]{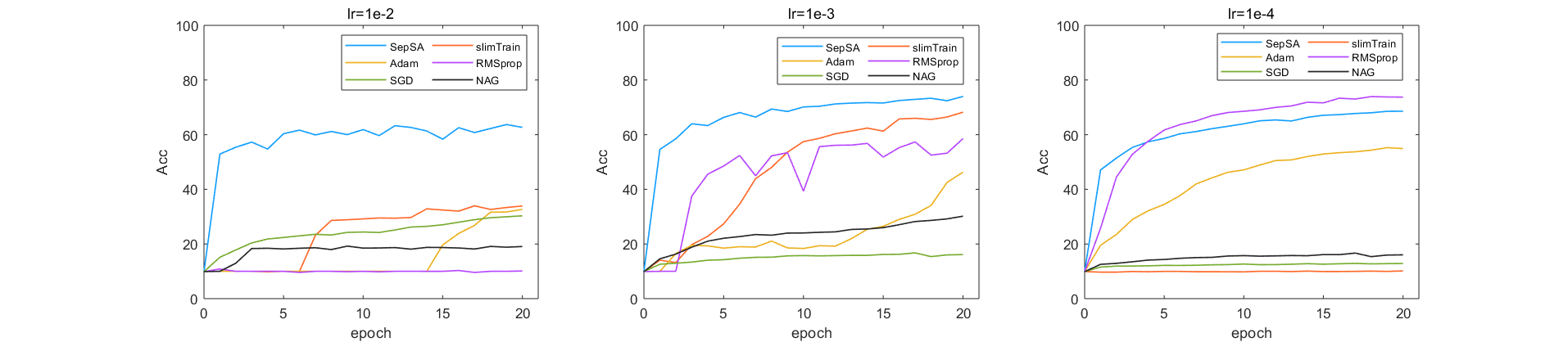}
	\noindent\colorbox{gray!20}{\makebox[\dimexpr0.81\linewidth-2\fboxsep-2\fboxrule]{CIFAR-10 test}}
	\includegraphics[width=1\linewidth]{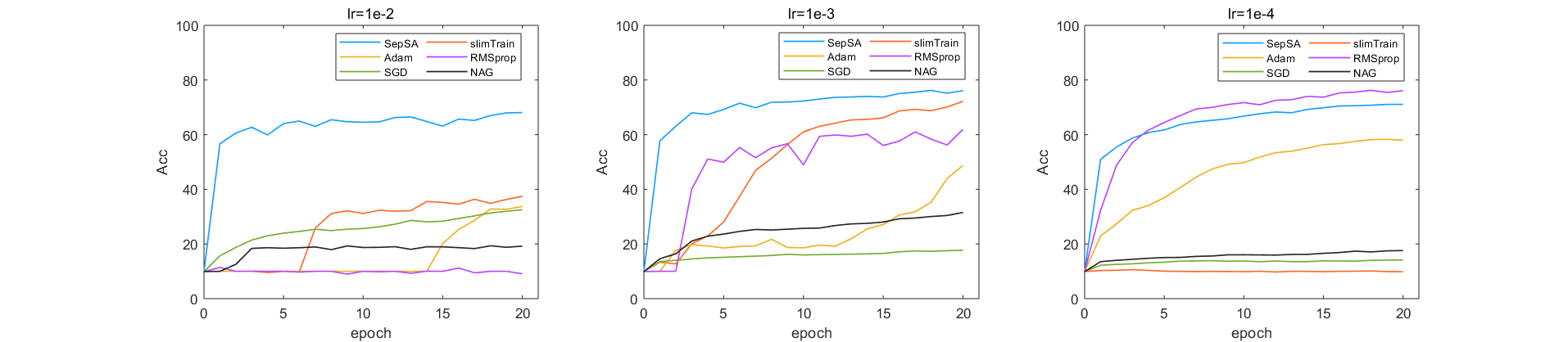}
	\caption{Mini-batch learning results of various algorithms on the classification datasets with different learning rates. The batch size is 64.}
\end{figure*}
\begin{figure*}[htp]
	\centering
	\includegraphics[width=1\linewidth, height=0.5\textwidth]{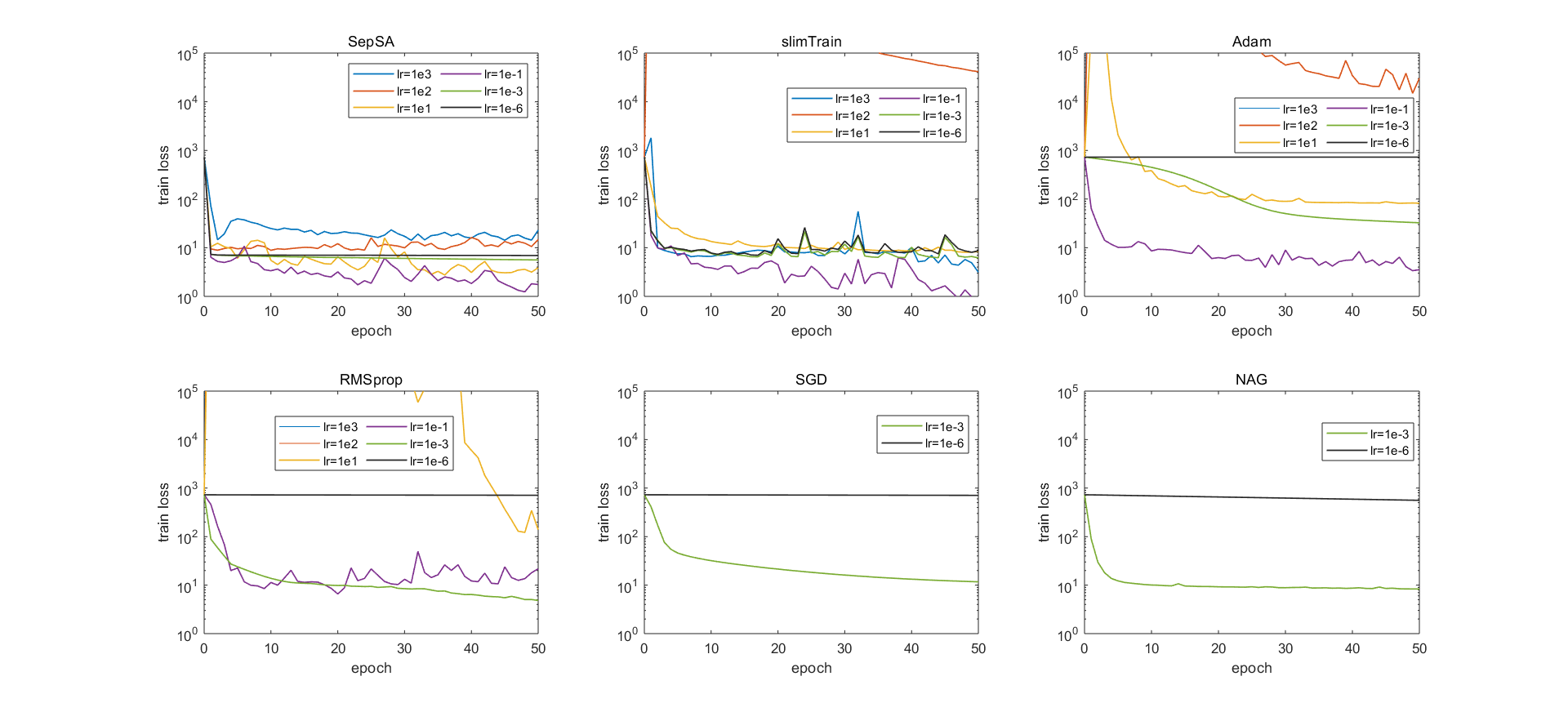}
	\caption{Results of different algorithms on the Energy efficiency dataset with a wide range of learning rates.}
\end{figure*}
\begin{table*}[h]
	\centering
	\caption{Statistical Results under Different Initial Values (100 different initial values for the regression and 10 different initial values for the classification). The number following $\pm$ is the deviation.}
	\begin{tabular}{|c|c|c|c|c|c|}\hline
		\makecell[c]{Type of \\Learning}& Dataset & Algorithm & Train (MSE or Accuracy) & Test (MSE or Accuracy) & Time (s)\\ \hline
		\hline
		\multirow{12}{*}{Online}
		&\multirow{3}{*}{Energy efficiency}
		& SepSA & \textbf{7.7356$\pm$0.3546} &	\textbf{8.3049$\pm$0.5443} & 0.3987$\pm$0.0198 \\ \cline{3-6}
		&& slimTrain & 1659.8121$\pm$3093.6973&	1847.5328$\pm3494.0649$&0.8427$\pm$0.0566	\\ \cline{3-6}
		&& Adam & 165.1144$\pm$21.6400 &	122.6856$\pm16.7853$& \textbf{0.3395$\pm$0.0193}\\ \cline{2-6}
		&\multirow{3}{*}{Diabetes} 
		& SepSA & \textbf{2481.3428$\pm$65.0569}&\textbf{2940.2236$\pm$129.6344}& 0.2420$\pm$0.0171\\ \cline{3-6}
		&& slimTrain &6943.6455$\pm$3215.9365&	6729.5498$\pm$2950.3054&0.5026$\pm$0.0443\\ \cline{3-6}
		&& Adam &26413.9551$\pm$262.0965&27033.5684$\pm$254.8461&\textbf{0.2362$\pm$0.0262}\\ \cline{2-6}
		&\multirow{3}{*}{MNIST}
		& SepSA &\textbf{98.40\%$\pm$0.14\%}&\textbf{98.39\%$\pm$0.15\%}&\textbf{519.2630$\pm$30.0794}\\ \cline{3-6}
		&& slimTrain &14.77\%$\pm$2.95\%&	14.74\%$\pm$3.08\%&	533.8472$\pm$19.3596\\ \cline{3-6}
		&& Adam &98.01\%$\pm$0.24\%&	97.97\%$\pm$0.29\%&535.8935$\pm$\textbf{13.3635}\\ \cline{2-6}
		&\multirow{3}{*}{CIFAR-10}
		& SepSA &\textbf{35.64\%$\pm$1.96\%}&\textbf{38.14\%$\pm$2.54\%}&\textbf{412.6066$\pm$13.0932}\\ \cline{3-6}
		&& slimTrain &10.17\%$\pm$0.68\%&	9.90\%$\pm$0.83\%&	442.2921$\pm$17.7732\\ \cline{3-6}
		&& Adam &22.61\%$\pm$1.54\%&	24.61\%$\pm$1.48\%&420.3238$\pm$12.4743 \\ \hline
		\hline
		\multirow{12}{*}{Mini-batch}
		&\multirow{3}{*}{Energy efficiency}
		& SepSA & \textbf{2.3869$\pm$0.3448}&\textbf{3.2776$\pm$0.5590}&0.8041$\pm$0.0302\\ \cline{3-6}
		&& slimTrain & 5.4708$\pm$1.3586 & 7.2606$\pm$1.7903 & 2.9818$\pm$0.1243\\ \cline{3-6}
		&& Adam & 29.9258$\pm$1.6203 & 18.3703$\pm$1.0336 & \textbf{0.6878$\pm$0.0341}\\ \cline{2-6}
		&\multirow{3}{*}{Diabetes} 
		& SepSA & \textbf{1291.8002$\pm$83.4968}&\textbf{3076.0276$\pm$171.3276}&0.4845$\pm$0.0252\\ \cline{3-6}
		&& slimTrain &1643.3928$\pm$153.9411&3733.7151$\pm$510.7436&1.7124$\pm$0.1257\\ \cline{3-6}
		&& Adam & 16805.7559$\pm$720.2620&17581.1328$\pm$726.8230&\textbf{0.4044$\pm$0.0214}\\ \cline{2-6}
		&\multirow{3}{*}{MNIST}
		& SepSA &\textbf{99.78\%$\pm$0.02\%}&	\textbf{99.23\%$\pm$0.03\%}&	184.8455$\pm$3.5047\\ \cline{3-6}
		&& slimTrain &99.48\%$\pm$0.15\%&	99.09\%$\pm$0.12\%&	196.8310$\pm$8.8517\\ \cline{3-6}
		&& Adam &99.69\%$\pm$0.04\%&	99.15\%$\pm$0.04\%&	\textbf{148.1506$\pm$5.0871}\\ \cline{2-6}
		&\multirow{3}{*}{CIFAR-10}
		& SepSA &\textbf{72.77\%$\pm$0.75\%}&	\textbf{74.95\%$\pm$0.59\%}&	448.5833$\pm$5.9535\\ \cline{3-6}
		&& slimTrain &66.79\%$\pm$1.35\%&	69.86\%$\pm$0.87\%&	458.6548$\pm$10.9953\\ \cline{3-6}
		&& Adam &43.30\%$\pm$5.60\%&	44.79\%$\pm$5.82\%&\textbf{383.8587$\pm$5.0369}\\ \hline
	\end{tabular}
\end{table*}
\subsection{Investigation of Learning Rate Sensitivity}
To further investigate the sensitivity of algorithms to different learning rates, we test a larger range of learning rates on the Energy Efficiency dataset and present the training error results in Fig. 4. Note that we excluded the results of previously tested learning rates ($1e-2$, $1e-4$) to avoid redundancy. Moreover, due to the occurrence of NAN values under larger learning rates, only the results of SGD and NAG algorithms under two specific learning rates are displayed. As shown in Fig. 4, SepSA is the least sensitive to the learning rate, maintaining a consistent decrease in training error for all tested learning rates, and exhibiting the smallest difference in training effect compared to other algorithms at varying learning rates. The second least sensitive algorithm is slimTrain, but it diverges when $lr=1e2$. Adaptive step size algorithms Adam and RMSprop follow as the next most sensitive and also diverge at $lr\geq 10$, with significant differences in results at various learning rates. Finally, SGD and NAG exhibit the highest sensitivity to the learning rate.

\subsection{Statistical Results under Different Initial Values}
Table III displays the statistical results of several algorithms across multiple initial values. Specifically, we use 100 different initial values for the regression datasets and 10 initial values for the classification datasets, as the latter required a longer training time. The widely-used Adam algorithm is chosen to represent SGD-like methods, and only results obtained with a learning rate of $1e-3$ are recorded.

The results in Table III demonstrate that the SepSA algorithm exhibits superior performance in both training and testing compared to other methods. In terms of running time, SepSA takes slightly longer than Adam, while slimTrain requires the most time among the three. Nevertheless, as SepSA has a faster initial convergence speed, it should require less training time than Adam to achieve the same level of loss or accuracy results.

\section{Conclusion}
In this paper, we have introduced a class of stochastic separable optimization problems and proposed an online learning algorithm for solving the stochastic separable nonlinear least squares problems under a separable stochastic approximation framework. The proposed algorithm focuses on optimizing models where some parameters have a linear nature, which is common in machine learning. The algorithm updates the linear parameters using the recursive least squares algorithm and then updates the nonlinear parameters using the stochastic gradient method. This can be understood as a stochastic approximation version of block coordinate descent approach. The global convergence of the proposed online algorithm for non-convex cases has been established in terms of the expected violation of a first-order optimality condition. Extensive experiments have been performed to compare the performance of the SepSA algorithm with other widely-used algorithms, and the experimental results demonstrate that the proposed algorithm exhibits notable advantages such as faster convergence speed, less sensitivity to learning rate, and more robust training and test performance. This paper provides a promising direction for solving separable stochastic optimization problems in machine learning and has practical implications for developing efficient online learning algorithms.

\vfill

\end{document}